%% file: article.tex
\documentclass[]{interact}

\usepackage{epstopdf}
\usepackage[caption=false]{subfig}

\usepackage[numbers,sort&compress]{natbib}
\bibpunct[, ]{[}{]}{,}{n}{,}{,}
\renewcommand\bibfont{\fontsize{10}{12}\selectfont}
\makeatletter
\def\NAT@def@citea{\def\@citea{\NAT@separator}}
\makeatother

\usepackage[hidelinks]{hyperref}
\theoremstyle{plain}

\theoremstyle{definition}

\theoremstyle{remark}

\usepackage{siunitx}
\usepackage{tikz}
\usepackage[european]{circuitikz}
\usepackage{pgfplots}
\usetikzlibrary{patterns, patterns.meta, shapes.geometric, backgrounds, fit, positioning}
\pgfplotsset{compat=1.18}
\usepackage{xcolor}

\definecolor{halfgray}{gray}{0.55}
\definecolor{webgreen}{rgb}{0,.5,0}
\definecolor{webbrown}{rgb}{.6,0,0}
\definecolor{Maroon}{cmyk}{0, 0.87, 0.68, 0.32}
\definecolor{RoyalBlue}{cmyk}{1, 0.50, 0, 0}
\definecolor{Black}{cmyk}{0, 0, 0, 0}

\begin{document}

\articletype{Research Article}

\title{Learning Dynamics in Memristor-Based Equilibrium Propagation}

\author{
\name{Michael D\"{o}ll\textsuperscript{a}\thanks{CONTACT Michael D\"{o}ll. Email: michael.doell@acm.org}, Andreas M\"{u}ller\textsuperscript{a} and Bernd Ulmann\textsuperscript{b,c}}
\affil{\textsuperscript{a}Department of Computer Science, Darmstadt University of Applied Sciences, Darmstadt, Germany; \textsuperscript{b}FOM University of Applied Sciences, Frankfurt am Main, Germany; \textsuperscript{c}anabrid, Frankfurt am Main, Germany}
}

\maketitle

\begin{abstract}
Memristor-based in-memory computing has emerged as a promising paradigm to overcome the constraints of the von Neumann bottleneck and the memory wall by enabling fully parallelisable and energy-efficient vector-matrix multiplications. We investigate the effect of nonlinear, memristor-driven weight updates on the convergence behaviour of neural networks trained with equilibrium propagation (EqProp). Six memristor models were characterised by their voltage-current hysteresis and integrated into the EBANA framework for evaluation on two benchmark classification tasks. EqProp can achieve robust convergence under nonlinear weight updates, provided that memristors exhibit a sufficiently wide resistance range of at least an order of magnitude.
\end{abstract}


\section{Introduction}

The von Neumann bottleneck, an inherent limitation of traditional von Neumann architectures~\cite{backus_can_1978}, and its broader manifestation, the memory wall, which also affects Harvard architectures~\cite{wulf_hitting_1995}, pose significant challenges in modern computing systems~\cite{kaurComprehensiveReviewProcessinginMemory2024}. 
Both arise from the physical separation of memory and processing units, where the limited memory bandwidth cannot match the significantly higher clock frequencies of contemporary processors. This imbalance leads to underutilised computational resources, resulting in severe performance constraints. Cache hierarchies merely shift the bottleneck rather than eliminating it~\cite{suri_caching_2020}. The situation is further exacerbated by the energy inefficiency of frequent data transfers between processors and memory, as the energy cost of data movement can be up to $100$ times higher than that of a single floating-point operation~\cite{mittalSurveyReRAMBasedArchitectures2018}. These challenges are particularly critical for neural networks due to the immense scale at which vector-matrix multiplications (VMMs), providing the networks' computational backbone, are performed during both training and inference~\cite{mittalSurveyReRAMBasedArchitectures2018, gholami_ai_2024}.

Given the escalating energy demands and performance bottlenecks of digital computing, analogue computing is re-emerging as a promising alternative paradigm, offering significant advantages in energy efficiency and computational performance by bypassing data shuttling between separate memory and processing units~\cite{ulmann_analog_2023}. Unlike digital devices, which execute stepwise instructions stored in memory, analogue systems rely on the formation of a model (an analogue) of a given problem and perform computations through direct connection of computing elements~\cite{ulmann_analog_2020, ulmann_analog_2023}. The absence of centralised memory enables perfect parallelism, where all elements operate simultaneously without synchronisation overhead or memory bottlenecks that constrain digital architectures.

Nevertheless, these advantages are accompanied by certain drawbacks~\cite{ulmann_analog_2023}. Analogue systems are limited in precision, typically achieving a resolution of three to four decimal places, and are constrained to solving problems whose size does not exceed their number of available computing elements. Since both paradigms possess complementary strengths, hybrid computing capitalises on their synergy~\cite{ulmann_analog_2020, ulmann_analog_2023}: In such systems, a digital computer and an analogue computer are tightly coupled. The digital computer provides storage and orchestrates high-level control, parameter management, and post-processing, whereas the analogue component acts as a co-processor that delivers the raw performance and efficiency necessary for specialised tasks.

In the context of neural networks, emerging in-memory computing architectures embody this paradigm shift.~\cite{kaurComprehensiveReviewProcessinginMemory2024}. In these systems, weight matrices are encoded as variable conductances in memristive devices, allowing VMMs to be performed directly in the analogue domain. Several prototypes already demonstrate this concept. Mythic, for example has developed an architecture that leverages flash-based analogue memory, where $8$-bit digital-to-analogue converters (DACs) program each cell to assume one of $256$ distinct conductance levels~\cite{demler_mythic_2018}. This allows the system to encode signed weight values ranging from $-127$ to $+127$. In a parallel approach, IBM employed phase-change memory (PCM) to realise a memristive implementation of synaptic weights~\cite{demler_ibm_2018}. In these devices, an electrical pulse heats a chalcogenide glass to initiate a transition from an amorphous to a crystalline state, modulating the conductance of each PCM cell.

Energy-efficient computing hardware not only reduces operational cost but also enables new applications~\cite{xue_recent_2016}. 
Mobile devices, constrained by limited battery size and weight, can benefit from extended operational times. Autonomous sensors could operate entirely on ambient energy harvesting, and implantable biomedical systems must minimise heat generation to protect surrounding tissue. However, achieving fully analogue neural networks requires training algorithms that can operate under the constraints and nonidealities of analogue circuits~\cite{aguirre_hardware_2024}. One promising approach to address this need is the equilibrium propagation (EqProp) framework~\cite{scellierEquilibriumPropagationBridging2017}. Nonetheless, nonlinear conductance changes characteristic of memristors~\cite{chuaMemristormissingCircuitElement1971} may influence the learning process.

Therefore, this work investigates how memristors influence the learning behaviour of nonlinear resistive networks (NRNs)~\cite{kendall_training_2020} trained with EqProp by examining how nonlinear conductance updates affect network convergence. Specifically, we identify which memristor models yield the most robust training performance, assess the effectiveness of different pulse-modulation schemes for regulating memristive weight updates, and determine the minimum ratio between maximum ($R_{\text{OFF}}$) and minimum ($R_{\text{ON}}$) device resistances required for stable learning.
To address these objectives, a simulation-based methodology was employed. Multiple memristor models were first characterised through analysis of their hysteresis curves and then integrated into an EqProp simulation framework to systematically evaluate their effect on NRN convergence.

\section{Background}

The memristor, proposed by Leon Chua in 1971 as a theoretical concept, was envisioned as a missing fundamental element in circuit theory~\cite{chuaMemristormissingCircuitElement1971}. Decades later, in 2008, researchers at Hewlett-Packard Labs provided experimental validation of its physical existence, developing the first functional solid-state implementation~\cite{strukovMissingMemristorFound2008}. Since then, memristors have gained significant attention for their potential in hardware acceleration of artificial neural networks, offering efficient, low-power alternatives to conventional computing architectures~\cite{aguirre_hardware_2024}.

The ideal memristor is defined by the relationship between electric charge $q$ and magnetic flux $\varphi$~\cite{chuaMemristormissingCircuitElement1971, valsaAnalogueModelMemristor2011}. Unlike conventional resistors with a fixed resistance, the resistance of the ideal memristor varies based on the charge (charge-controlled) or flux (flux-controlled). These quantities are linked to the time-domain behaviours of current $i(t)$ and voltage $u(t)$ through the following integrals, which enables the memristor to function as a nonlinear resistor with memory, capable of retaining states.
\begin{equation}
q(t) = \int_{-\infty}^{t} i(\tau) \, \mathrm{d}\tau, \quad
\varphi(t) = \int_{-\infty}^{t} u(\tau) \, \mathrm{d}\tau
\label{eq:integrals}
\end{equation}

In a charge-controlled memristor, the magnetic flux depends on the charge, which itself is controlled by the history of current~\cite{chuaMemristormissingCircuitElement1971, valsaAnalogueModelMemristor2011}. The memristance $M(q)$, representing the resistance $R$ at a specific state determined by the charge, is defined by the derivative of flux with respect to charge. Similarly, in a flux-controlled memristor, the charge is determined by the magnetic flux and its dependence on the history of voltage. The memductance $W(\varphi)$, which is the inverse of the memristance, describes how the conductance $G$ varies with flux.
\begin{equation}
M(q) = \frac{\mathrm{d}\varphi(q)}{\mathrm{d}q}, \quad
W(\varphi) = \frac{\mathrm{d}q(\varphi)}{\mathrm{d}\varphi}
\end{equation}

A key characteristic of memristors is their pinched hysteresis loop in the current-voltage relationship, which varies significantly with the frequency and waveform of the input signal~\cite{strukovMissingMemristorFound2008}.

\subsection{Hardware Acceleration of Vector-Matrix Multiplication}
\label{sec:02:01}

Memristor-based architectures overcome the limitations of the von Neumann bottleneck and the memory wall by integrating memory and computation into a single unit, thereby reducing data movement, mitigating energy inefficiencies, and enhancing computational throughput~\cite{baoMemristiveInmemoryComputing2022}. To leverage these advantages in artificial neural networks, memristors are explored for efficiently implementing VMM directly in hardware~\cite{szeEfficientProcessingDeep2017}. Every VMM computation can be decomposed into a series of multiply-accumulate (MAC) operations, where every MAC involves multiplying an input value by a corresponding weight and aggregating its product to compute an output:
\begin{equation} a \leftarrow a \cdot w + b \end{equation}

MAC operations can be implemented using memristors by exploiting Ohm's law and Kirchhoff's current law (KCL)~\cite{szeEfficientProcessingDeep2017, baoMemristiveInmemoryComputing2022}. If the conductance $G$ represents the weight and the applied voltage U corresponds to the input value, the multiplication of the input by the weight is inherently performed through Ohm's law $I = G \cdot U$, where $I$ is the current flowing through the memristor~\cite{aguirre_hardware_2024}. To perform the accumulate step, KCL is applied, which dictates that the total current entering a node is equal to the total current leaving it~\cite{lepri_-memory_2023, szeEfficientProcessingDeep2017}. For multiple memristors connected to a common output node, the individual currents generated by each memristor are summed at the node, with the resulting total current representing the accumulated weighted inputs $\smash{I_{\text{out}} = \sum U_i \cdot G_i}$, where $G_i$ denotes the conductance of the $i$-th memristor, and $U_i$ represents its corresponding input voltage. Based on this principle, crossbar arrays can be constructed to efficiently perform VMM in hardware~\cite{liuMemristorCrossbarBased2016, aguirre_hardware_2024}. In a crossbar array, memristors are positioned at the intersections of horizontal conductive lines (referred to as wordlines) and vertical conductive lines (referred to as bitlines). This configuration creates a gridlike structure where each memristor at position $(i,j)$ encodes a weight $G_{i,j}$. The input voltages $U_i$ corresponding to the elements of the input vector are applied along the wordlines, and the resulting output currents $I_j$, representing the accumulated weighted sums, are collected along the bitlines. This architecture, illustrated in Figure~\ref{fig:crossbar-array} and described by Equation \ref{eq:vmm}, enables the simultaneous computation of all weighted sums in $\mathcal{O}(1)$ time complexity, using the physical properties of memristors to achieve a highly parallel and energy-efficient VMM~\cite{baoMemristiveInmemoryComputing2022}. Memristor crossbar architectures may also enable $3$D stacking, allowing multiple crossbar layers to be integrated vertically to increase computational density and perform larger-scale computations within the same area~\cite{liThreedimensionalCrossbarArrays2017, aljafar3DCrossbarArchitecture2021}.
\begin{equation}
\begin{aligned}
\begin{bmatrix}
U_1 & U_2 & \dots & U_i
\end{bmatrix}
\cdot
\begin{bmatrix}
G_{1,1} & G_{1,2} & \dots & G_{1,j} \\
G_{2,1} & G_{2,2} & \dots & G_{2,j} \\
\vdots & \vdots & \ddots & \vdots \\
G_{i,1} & G_{i,2} & \dots & G_{i,j}
\end{bmatrix}
=
\begin{bmatrix}
I_1 & I_2 & \dots & I_j
\end{bmatrix}
\label{eq:vmm}
\end{aligned}
\end{equation}
\input{figures/crossbar-array}

To incorporate a bias term into the VMM operation, an additional wordline can be introduced that applies a voltage $U_B=\text{const.}$ serving as the bias input~\cite{aguirre_hardware_2024}. An inherent limitation of memristor crossbar arrays lies in their inability to represent negative conductance values~\cite{aguirre_hardware_2024,kendall_training_2020}. As a result, the implementation of negative weights is not directly feasible within the array, also several methods exist to address this constraint.
One approach involves duplicating the wordlines~\cite{aguirre_hardware_2024,kendall_training_2020} as illustrated in Figure~\ref{fig:input-duplication}. For each input voltage $U_i$, an additional input node with the inverted voltage $-U_i$ is introduced. The conductance matrix is expanded to represent the positive and negative weights separately. Specifically, the conductances $G_{2i-1}$ encode the positive and $G_{2i,j}$ the negative contributions of the weights.
In another approach the bitlines are duplicated~\cite{aguirre_hardware_2024,kendall_training_2020} as depicted in Figure~\ref{fig:output-duplication}. Here, the input nodes remain unchanged, while the conductance matrix is expanded to include separate columns for positive and negative weights. Each original output node $I_j$ is split into two separate output nodes: $\smash{I_j^+}$ (representing the positive contributions) and $\smash{I_j^-}$ (representing the negative contributions). To compute the final output, the difference between the duplicated outputs is calculated, which can be implemented in hardware using a transimpedance amplifier to convert a current into a voltage, followed by a differential amplifier to perform the final subtraction.
\input{figures/line-duplication}

\subsection{Equilibrium Propagation}
\label{sec:chapter02:section03}

The framework of EqProp offers a more biologically plausible and hardware-efficient alternative to backpropagation, particularly suitable for implementation in memristor-based neuromorphic hardware~\cite{scellierEquilibriumPropagationBridging2017}. Unlike backpropagation, which requires separate computational pathways for the forward and backward passes, EqProp uses a single set of neural dynamics for both inference and learning.

\subsubsection{Theoretical Framework}

EqProp is designed to train energy-based models (EBMs), a class of machine-learning models that encode dependencies between variables through an energy function~\cite{scellierEquilibriumPropagationBridging2017, watfaEnergybasedAnalogNeural2023}. This energy function $E(\theta, x, s)$ is a scalar function of parameters $\theta$, input $x$, and network state $s$, which assigns lower values to desired states and higher values to undesirable state configurations~\cite{scellier_agnostic_2022, scellier_energy-based_2023}. Given fixed $\theta$ and $x$, the model relaxes into an equilibrium state of minimal energy:
\begin{equation}
s(\theta, x) = \underset{s}{\arg\min} \; E(\theta, x, s)
\end{equation}

The learning process in EqProp is divided into two different phases: the free phase and the nudging phase~\cite{scellierEquilibriumPropagationBridging2017, watfaEnergybasedAnalogNeural2023}. In the free phase, the inputs are provided, and the system settles into a free equilibrium state $s_0$. This corresponds to the network's inference process, yielding the model's predictions at the output units. In the nudging phase, an error is introduced at the output units, nudging them towards the target values. This causes the network to converge to a weakly clamped equilibrium state $s_{\beta}$ with a lower prediction error~\cite{scellierEquilibriumPropagationBridging2017, scellier_agnostic_2022}:
\begin{equation}
s_{\beta}(\theta, x, y) = \underset{s}{\arg\min} \left[ E(\theta, x, s) + \beta \, C(s, y) \right]
\label{eq:nudged-phase}
\end{equation}

In this phase, the total energy is partitioned into two summands: the previously defined internal energy $E(\theta, x, s)$, which models the interactions among the network's components, and the external energy $C(s, y)$, representing the cost function that captures the discrepancy between the predicted output (the state at the output nodes) and the desired target $y$~\cite{scellierEquilibriumPropagationBridging2017, watfaEnergybasedAnalogNeural2023, scellier_energy-based_2023}. The hyperparameter $\beta \in \mathbb{R}$ controls the influence of the external energy. During the free phase $\beta = 0$ causing the system to minimise only the internal energy. In the nudging phase $\beta \neq 0$ so that the system is nudged towards the target values by minimising the total energy. The parameter updates are performed via gradient descent $\smash{\Delta\theta = -\eta \frac{\partial \mathcal{L}}{\partial \theta}(\theta, x, y)}$~\cite{scellier_energy-based_2023} with learning rate $\eta$ and loss function $\smash{\mathcal{L}(\theta, x, y) = C(s(\theta, x), y)}$~\cite{scellier_agnostic_2022}. The loss gradient can be calculated by analysing how the gradient of the system's energy with respect to the model parameters changes when applying a small perturbation towards the target~\cite{scellier_agnostic_2022}:
\begin{equation}
\frac{\partial \mathcal{L}}{\partial \theta}(\theta, x, y) = \left. \frac{d}{d\beta} \right|_{\beta=0} \frac{\partial E}{\partial \theta}(\theta, x, s_\beta)
\end{equation}
\input{figures/eqprop-flowchart}

Since the exact gradient requires an infinitesimally small value of $\beta$, a surrogate loss function $\mathcal{L}_{\beta}$ must be introduced to approximate the true loss $\mathcal{L}$ and allow for practical implementation of the gradient descent step by using the first-order finite difference forward estimator~\cite{scellier_agnostic_2022, scellier_energy-based_2023}:
\begin{equation}
\frac{\partial \mathcal{L}_{\beta}}{\partial \theta}(\theta, x, y) = \frac{1}{\beta} \left( \frac{\partial E}{\partial \theta}(\theta, x, s_\beta) - \frac{\partial E}{\partial \theta}(\theta, x, s_0) \right)
\end{equation}

Thus, the general procedure of EqProp can be summarised in the following steps, as depicted in Figure~\ref{fig:eqprop-flowchart}~\cite{scellierEquilibriumPropagationBridging2017,watfaEnergybasedAnalogNeural2023}:
\begin{enumerate}
    \item \textbf{Free phase}: Fix the input $x$ at the input nodes and allow the system to settle to the free equilibrium state $s_0$. Collect $\smash{\frac{\partial E}{\partial \theta}}$ at this state.
    \item \textbf{Nudging phase}: Apply a nudging signal to the output nodes, pushing them weakly towards their target values. Let the system relax to the nudged equilibrium state $s_\beta$. Again, collect $\smash{\frac{\partial E}{\partial \theta}}$ at this new state.
    \item \textbf{Parameter update}: Compute the difference of these gradients, scaled by $\smash{\frac{1}{\beta}}$ to calculate $\smash{\frac{\partial \mathcal{L}_{\beta}}{\partial \theta}}$, and use this to update the parameters via gradient descent.
\end{enumerate}

This cycle is repeated for each training sample, with parameter updates performed based on the chosen batching strategy (per sample, mini-batch, or batch)~\cite{scellierEquilibriumPropagationBridging2017, ruder_overview_2017}. The sign of the nudging parameter $\beta$ distinguishes two variants of the EqProp learning rule. For $\beta > 0$, the procedure is known as positively perturbed EqProp, which provides a lower bound of the cost function~\cite{scellier_energy-based_2023}. Conversely, when $\beta < 0$, it is termed negatively perturbed EqProp, resulting in an upper bound on the cost function:
\begin{equation}
\mathcal{L}_{\beta}(\theta, x, y) \leq C(s(\theta, x), y) \leq \mathcal{L}_{-\beta}(\theta, x, y), \quad \forall \beta > 0
\end{equation}

For $\beta \to 0$, the function $\mathcal{L}_{\beta}$ approximates $C$ up to $\mathcal{O}(\beta)$~\cite{scellier_energy-based_2023}. To further reduce this error, a second nudging phase can be performed using the opposite sign for the nudging factor ($-\beta$), an approach referred to as centred EqProp, where the following symmetric difference estimator is used to compute the loss gradient~\cite{laborieux_scaling_2021, scellier_energy-based_2023}:
\begin{equation}
\frac{\partial \mathcal{L}_{-\beta;+\beta}}{\partial \theta}(\theta, x, y) = \frac{1}{2\beta} \left( \frac{\partial E}{\partial \theta}(\theta, x, s_{-\beta}) - \frac{\partial E}{\partial \theta}(\theta, x, s_\beta) \right)
\end{equation}

With centred EqProp, the surrogate loss $\mathcal{L}_{-\beta;+\beta}$ constitutes a second-order approximation of $C$~\cite{scellier_energy-based_2023}:
\begin{equation}
\mathcal{L}_{-\beta;+\beta}(\theta, x, y) = C(s(\theta, x), y) + \mathcal{O}(\beta^2)
\end{equation}

\subsubsection{Proposed Hardware Implementation}

EqProp seamlessly aligns with NRNs, which are a subset of EBMs~\cite{kendall_training_2020}. NRNs consist of analogue electrical circuits composed of two-terminal components, including memristors, diodes, voltage sources, and current sources. In an NRN, the internal energy is based on the architecture of the circuit and arises from the electrical dynamics of the components so that it is naturally minimised by the governing physical laws.

A significant advantage of EqProp compared with backpropagation lies in its hardware efficiency~\cite{watfaEnergybasedAnalogNeural2023}. A major challenge for efficient hardware implementations of backpropagation using memristor crossbars is the necessity for separate forward and backward passes, each typically requiring distinct circuit configurations~\cite{kendall_training_2020}. Due to unavoidable device variations and nonidealities inherent in analogue circuits, systematic gradient errors arise, which accumulate throughout backward propagation of gradients through the network, leading to significant degradation in learning performance. To counteract these errors, analogue-to-digital converters (ADCs) and DACs are employed between layers, allowing for precise calculation of activations and their derivatives in the digital domain. However, these conversions not only introduce considerable hardware complexity but also result in substantial energy consumption, as ADCs are the largest and most power-hungry components in such systems~\cite{aguirre_hardware_2024}. In contrast, EqProp eliminates the need for separate circuits dedicated to forward and backward passes and utilises a unified hardware architecture for both free and nudging phases, simplifying the design and enabling fully analogue computation across both phases~\cite{kendall_training_2020}. Additionally, the weight updates are inherently driven by local changes in energy, as the energy function in NRNs is entirely sum-separable~\cite{watfaEnergybasedAnalogNeural2023}.

As NRNs utilise memristor crossbar arrays for weight storage and VMMs, which can be integrated into the EqProp framework using the implementation described in section \ref{sec:02:01}, the outlined approaches to represent negative weights may also be implemented by doubling the input nodes (resulting in duplicated wordlines in the first layer) and the output nodes (leading to duplicated bitlines in the last layer) of the model~\cite{kendall_training_2020}. However, representing negative weights in hidden layers requires alternative methods, as EqProp demands the NRN to be fully bidirectional.
\input{figures/non-linearity}

Nonlinearities can be implemented using diode-based circuits, enabling the conversion of currents into voltages in a nonlinear manner~\cite{kendall_training_2020}. For example, an activation function similar to a rectified linear unit (ReLU) can be realised by placing a diode in series with a direct current (DC) voltage source~\cite{watfaEnergybasedAnalogNeural2023}. The breakpoint of the ReLU is adjustable through the amount of voltage provided by the DC source. Since the voltage also drops slightly beyond the breakpoint, the ReLU exhibits a small leakage. By connecting two diodes in opposite orientations in parallel and pairing each with its own voltage source, it is possible to create two breakpoints, allowing the construction of a sigmoid-like profile~\cite{kendall_training_2020}. The circuit diagrams of these nonlinearities are illustrated in Figures~\ref{fig:relu-circuit} and~\ref{fig:sigmoid-circuit}, and their corresponding $U$-$I$ characteristics in Figures~\ref{fig:relu} and~\ref{fig:sigmoid}. In a neuron, a nonlinearity is complemented by a bidirectional amplifier to prevent signal attenuation in both directions. A bidirectional amplifier combines a voltage-controlled voltage source (VCVS) and a current-controlled current source (CCCS), as shown in Figure~\ref{fig:bidirectional-amplifier}. The VCVS amplifies voltages in the forward direction with a gain $A$, which is a hyperparameter, to prevent voltage decay as the signal traverses layers, while the CCCS reduces currents in the backward direction with an attenuation of $\frac{1}{A}$.
\input{figures/nrn-architecture}

An example of a layered NRN architecture is depicted in Figure \ref{fig:nrn-architecture}~\cite{kirazImpactsFeedbackCurrent2022}. The network consists of input nodes connected to DC voltage sources, hidden neurons, and output nodes, all integrated through memristor crossbar arrays. To represent negative weights while maintaining full bidirectionality, in the first crossbar array the wordlines and in the second crossbar array the bitlines are duplicated.

During the free phase, the DC input sources encode the input data of the network~\cite{kendall_training_2020}. The first crossbar array performs a VMM, and the resulting currents are fed into the hidden neurons. Subsequently, the second crossbar array computes another VMM, whose output voltages represent the predictions of the model after reaching equilibrium. $\hat{Y}_k^{+,0}$ and $\hat{Y}_k^{-,0}$ denote the free-phase predictions for the positive and negative components, respectively. In the nudging phase, feedback currents proportional to the gradient of the loss function, scaled by the nudging factor $\beta$, are injected into the output nodes. As a result, the system shifts to a new equilibrium state closer to the target output $Y_k$. If mean squared error (MSE)~\cite{james_introduction_2023} is used as cost function, the feedback currents are defined as follows~\cite{kendall_training_2020}:
\begin{equation}
I_k^+ = \beta \left( Y_k - \hat{Y}_k^{+,0} + \hat{Y}_k^{-,0} \right),\quad I_k^- = \beta \left( \hat{Y}_k^{+,0} - \hat{Y}_k^{-,0} - Y_k \right)
\end{equation}

The voltage drops $\smash{\Delta U_{ij}}$ across each memristor during the free and nudged phases serve as the foundation for the local gradient approximation required for updating the weights~\cite{kendall_training_2020}. For a device at crosspoint $(i,j)$, the input and output voltages are $\smash{U_i^0}$ and $\smash{U_j^0}$ in the free equilibrium state as well as $\smash{U_i^\beta}$ and $\smash{U_j^\beta}$ in the nudged equilibrium state:
\begin{equation}
\Delta U_{ij}^0=U_i^0-U_j^0,\quad \Delta U_{ij}^\beta=U_i^\beta-U_j^\beta
\end{equation}

The difference between the squared voltage drops in the two phases provides the approximation of the gradient~\cite{kendall_training_2020}. These are scaled by the learning rate $\eta$ and the nudging factor $\beta$ so that the conductances $G_{ij}$ are updated as follows during stochastic gradient descent:
\begin{equation}
G_{ij}\leftarrow G_{ij}-\frac{\eta}{\beta}\left[\left(\Delta U_{ij}^\beta\right)^2-\left(\Delta U_{ij}^0\right)^2\right]
\end{equation}

\section{Methods}

Since memristors exhibit nonlinear conductance changes that influence weight updates~\cite{chuaMemristormissingCircuitElement1971}, modelling of memristors is essential to incorporate these nonlinearities into simulations of the analogue EqProp approach. Six memristor models were implemented in Python to evaluate their current-voltage hysteresis behaviour:
\begin{itemize}
    \item Linear ion drift model~\cite{strukovMissingMemristorFound2008}
    \item Joglekar model~\cite{joglekarElusiveMemristorProperties2009}
    \item Biolek model~\cite{biolekzdenekSPICEModelMemristor2009}
    \item VTEAM model~\cite{kvatinskyVTEAMGeneralModel2015}
    \item Yakopcic model~\cite{yakopcic_memristor_2011}
    \item MMS model~\cite{molter_mean_2017}
\end{itemize}

Complementing these memristor models, a baseline model exhibiting linear conductance changes was implemented to serve as a reference for comparison. Each model was driven by sinusoidal voltage waves with an amplitude of $\SI{1}{\volt}$ at empirically selected frequencies that best revealed its switching dynamics. Lower frequencies allow the internal state to evolve faster, while higher frequencies demonstrate the typical collapse of the hysteresis loop~\cite{vourkasMemristorFundamentals2016}. When the frequency is sufficiently low, the device reaches both its low-resistance state (LRS) and high-resistance state (HRS) within exactly half a sine period. Further increasing the pulse width would provide no additional benefit since the remaining portion of the waveform no longer contributes to the state transition. Using identical frequencies across all models would have obscured key features of certain models due to either excessively compressed or prematurely saturated hysteresis loops.

To investigate the impact of memristors on EqProp convergence, simulations were performed using the EBANA framework~\cite{watfaEnergybasedAnalogNeural2023}. EBANA implements EqProp by combining SPICE simulations of network dynamics during the free and nudged phases with Python-based gradient approximations and weight updates. The framework was extended to incorporate the implemented memristor models for weight updates. The baseline model with linear conductance changes was also incorporated during training to enable direct comparison of linear weight updates and nonlinear weight updates incorporating intrinsic memristive characteristics.

\subsection{Datasets and Network Architecture}
Given the computational constraints of the SPICE simulations within EBANA, the following two simple classification datasets were chosen: \begin{itemize}
    \item \textbf{Iris Flowers}~\cite{r.a.fisherIris1936}: Consists of $150$ samples, each with $4$ input features (sepal length, sepal width, petal length, petal width). The dataset is split into $3$ classes representing different iris species.
    \item \textbf{Breast Cancer Wisconsin (Diagnostic)}~\cite{williamwolbergBreastCancerWisconsin1993}: Contains $569$ samples, each with $30$ input features. These features were computed from digitised images of fine-needle aspirates of breast masses. The binary output indicates whether the tumour is malignant or benign.
\end{itemize}

Each network is comprised of three layers of neurons (an input layer, one hidden layer, and an output layer) interconnected by two memristive crossbar arrays. Negative weights were implemented via wordline or bitline duplication, doubling the number of neurons in the input and output layers, with an additional bias node incorporated into the input layer. For the iris dataset, the network was trained using hidden layers containing $10$, $5$, and $2$ neurons, whereas for the breast cancer dataset, hidden layers of $16$, $10$, and $5$ neurons were evaluated. These configurations were chosen to investigate whether varying the network size would lead to different convergence behaviours for each dataset across the considered memristor models.

To investigate the effect of device-level parameters on learning, the ratio $\smash{\frac{R_{\text{OFF}}}{R_{\text{ON}}}}$ was varied in each experiment. A constant $R_{\text{ON}}$ of $\SI{100}{\ohm}$ was maintained, while $R_{\text{OFF}}$ was assigned multiple values (\SI{500}{\ohm}, \SI{1}{\kilo\ohm}, \SI{10}{\kilo\ohm}, and \SI{100}{\kilo\ohm}). This systematic variation allowed for an exploration of how the dynamic range of memristance impacts model convergence in EqProp. Convergence was assessed by monitoring the loss over $50$ epochs of training. In all experiments, the internal state variables of the memristors were randomly initialised using a uniform distribution of memristances. Moreover, for each combination of $R_{\text{OFF}}$, number of hidden neurons, and dataset, the same initial weights were used to ensure that differences in the observed convergence behaviour can be attributed to the memristor models and not to differences in the initial weights.

The learning rates were empirically determined by evaluating loss curves and selecting the values that yielded the most favourable convergence behaviour. The nudging factor $\beta$ was fixed at $1\times10^{-6}$ (positively perturbed EqProp). MSE~\cite{james_introduction_2023} served as the cost function and the gain factor was set to $A=4$. During preprocessing, all input features were normalised to the range of $[-0.5\,\text{V}, 0.5\,\text{V}]$ and the bias input was maintained at a constant voltage of \SI{0.5}{\volt} to reflect the practical voltage ranges used during SPICE simulations. Since the goal was to test basic viability and identify where convergence might fail, no validation or test datasets were included. The entire datasets were employed as training datasets with full-batch gradient descent.

\subsection{Pulse-Based Weight-Update Strategies}

During training, weight updates were executed by applying rectangular voltage pulses. For each weight update, a single voltage pulse was applied to modify the state of the corresponding memristor. Furthermore, memristance was exclusively modified during the weight update phase, while it remained fixed during the free and nudged phases. This approach implicitly assumes the existence of a voltage threshold that must be exceeded to induce changes in memristance.
\input{figures/modulation-schemes}

Two modulation schemes, namely pulse width modulation (PWM) and pulse amplitude modulation (PAM)~\cite{tomasi_advanced_2014} were implemented to update the state of each memristor in proportion to the gradient. In PWM, the voltage amplitude is fixed while the pulse width varies according to the gradient magnitude. In contrast, PAM applies pulses of fixed width with an amplitude proportional to the gradient magnitude. In the experiments, a constant amplitude of \SI{1}{\volt} was applied under the PWM scheme. For the PAM approach, the pulse duration was individually chosen for each memristor model, using specific pulse frequencies that demonstrated particularly favourable hysteresis characteristics. An illustration comparing the PWM and PAM schemes is provided in Figure~\ref{fig:modulation-schemes}.

An Adam optimiser~\cite{kingma_adam_2017} was employed to compute the weight updates due to its intrinsic ability to scale the learning rates for individual parameters. This characteristic permits effective convergence even under suboptimal learning rate settings. The exponential decay rates for the first- and second-moment estimates were set to the default values in EBANA of $0.5$. After computing gradient-based update values with the Adam optimiser for each parameter, the update signals were modulated via either PWM or PAM to control the voltage pulse applied to each memristor.
\input{figures/modulation-updates}

In the implementation of the PWM scheme as shown in Figure~\ref{fig:pwm-updates} the polarity of the applied voltage pulse was selected according to the sign of the update value. Specifically, a positive update value resulted in a negative voltage pulse; conversely, a negative update value corresponded to a positive voltage pulse. This sign inversion ensured that the memristor state moved in the direction opposite to the gradient. The pulse duration represented the absolute magnitude of the update, normalised by the fixed voltage amplitude to reduce sensitivity of the effective learning rate from future changes in operating voltage. Because the hysteresis of the VTEAM model is reversed with respect to the other memristor models, the polarity of the applied voltage was inverted before the state update when using the VTEAM model. Finally, the state of each memristor was updated based on voltage amplitude and pulse duration. In the PAM implementation (Figure~\ref{fig:pam-updates}), the pulse duration was fixed to the reciprocal of the operating frequency. The applied voltage was derived by scaling the negative update value by the pulse duration to minimise any effects of frequency changes on the effective learning rate. As with PWM, the amplitude was inverted for the VTEAM model prior to updating the memristor state.

\subsection{Considerations for Yakopcic Model Integration}

To integrate the Yakopcic memristor model into the EqProp simulation framework, several adaptations were required. In its original formulation, the Yakopcic model defines two distinct scaling factors, $a_1$ and $a_2$, for positive and negative voltages, respectively~\cite{yakopcic_memristor_2011}. However, our implementation in the EBANA framework only supported memristor models whose memristance remains invariant with respect to the polarity of the applied voltage. Therefore, the model was simplified by introducing a unified scaling factor $a = a_1 = a_2$, yielding
\begin{equation}
    M(t) = \frac{1}{a\, x(t)\, \sinh(b)} .
\end{equation}

For small $b$, the nonlinearity of the $\sinh(b)$ term becomes negligible, producing an almost linear current-voltage relation. Figure~\ref{fig:yakopcic-current} illustrates the resulting current-voltage characteristic for $b=0.05$ and $a=0.2$, ranging from $\SI{-1}{\volt}$ to $\SI{+1}{\volt}$. As shown, the nonlinearity is barely noticeable under these settings.
\input{figures/yakopcic-current}

Because the EqProp simulation framework required a unique, well-defined memristance value, the memristance was consistently evaluated at $\SI{\pm1}{\volt}$. To ensure that the simulated device exhibited finite resistance limits, the state variable $x$ was bounded between $x_\text{on}$ and 1, corresponding to $R_\text{ON}$ and $R_\text{OFF}$. Additionally, the window function for negative input voltages, $f_n(x)$, was modified to ensure it smoothly transitions to zero at $x_\text{on}$, leading to the following expression:
\begin{equation}
    f_n(x(t)) =
    \begin{cases}
    0, & \text{if}\;x(t) < x_\text{on} \\
    \exp\left[\alpha_n (x(t) + x_n - 1)\right]\,\frac{x(t) - x_\text{on}}{x_n - x_\text{on}}, & \text{if}\;x_\text{on} \le x(t) \le 1 - x_n \\
    1, & \text{otherwise}
    \end{cases}
\end{equation}

\section{Results}

We present hysteresis characteristics of the six memristor models at two drive frequencies and their respective impact on EqProp convergence across resistance ranges, hidden layer sizes, datasets, and pulse-based update schemes, summarised by the minimal loss during training.

\subsection{Characterisation of Memristor Models}

Figure~\ref{fig:hysteresis-comparison} compares the current-voltage hysteresis loops at two operating frequencies for each memristor model. Across models, increasing frequency compresses the loop as the the internal state has less time to evolve.

The linear ion drift model and the Joglekar model both exhibit point-symmetric loops with frequency-dependent widening. The Biolek model shows a clear asymmetry at the lower frequency, along with a slight drift. This drift becomes more pronounced at the higher frequency. The threshold-based models (VTEAM, Yakopcic, and MMS) manifestly evince their threshold behaviour, with state transitions occurring upon crossing their voltage thresholds. At lower frequencies, these models exhibit sharp transitions, whereas at higher frequencies, the transitions smooth out into more continuous loops.
\input{figures/hysteresis-comparison}

Unlike the other models, the VTEAM model exhibits a hysteresis loop that evolves in the opposite direction: from LRS to HRS during the positive half-wave, and vice versa. Furthermore, the VTEAM model exhibits rapid switching from HRS to LRS as soon as the threshold voltage is reached during the negative half-wave.

\subsection{Simulation of Equilibrium Propagation}

The minimal losses obtained across all configurations during EqProp training are illustrated in the heatmaps in Figures~\ref{fig:heatmap-pwm} and~\ref{fig:heatmap-pam}, corresponding to PWM and PAM, respectively. Each panel represents the combination of a specific dataset and hidden layer size, while the columns indicate the maximum resistance $R_\text{OFF}$ and the rows the memristor model. Green cells represent lower minimal losses and therefore good convergence, whereas red tones correspond to poor learning performance.
\input{figures/heatmap-pwm}
\input{figures/heatmap-pam}

Under PWM, EqProp achieved stable convergence when $R_\text{OFF} \ge \SI{1}{\kilo\ohm}$. At $R_\text{OFF}=\SI{500}{\ohm}$, all networks failed to converge, yielding persistently high loss values. Models trained on the breast cancer dataset generally exhibited higher minimal loss values than those trained on the iris dataset, reflecting its greater task complexity. Nonlinear memristor models performed comparably to the linear baseline.

PAM-based training revealed similar overall behaviour. Again, effective learning occurred for $R_\text{OFF} \ge \SI{1}{\kilo\ohm}$. For the VTEAM and Yakopcic models it was not feasible to utilise the same operating frequencies as under PWM due to their threshold-driven behaviour, as voltages near the switching thresholds caused excessive state changes. To stabilise learning, higher excitation frequencies were applied (\SI{5}{\giga\hertz} for VTEAM and \SI{300}{\kilo\hertz} for Yakopcic). Compared to the linear baseline, the MMS model exhibited higher losses for networks with smaller hidden layer sizes on the breast cancer dataset.

For models whose state change depends only on the voltage-time integral (linear updates, linear ion drift, Joglekar, and Biolek), PWM and PAM yielded identical learning curves when the effective $\int U\,\mathrm{d}t$ was matched between modulation schemes, with the learning rate $\eta$ and nudging factor $\beta$ constant. Across both modulation approaches EqProp learning proved robust, provided that a sufficient dynamic range was available ($\smash{\frac{R_\text{OFF}}{R_\text{ON}} \ge 10}$). Performance degraded sharply when this ratio fell below that threshold, independent of model, hidden layer size, or dataset.

\section{Discussion}

Our results indicate that convergence is governed primarily by the available resistance range rather than the specific memristor model. When this range narrows, weights press against the bounds of the realisable interval: the network can no longer represent the weight values required to further reduce the loss, and the weight updates are effectively clipped at the device limits. Model-specific effects matter mainly through their interaction with drive conditions. In particular, thresholded devices require a combination of excitation frequencies and voltages that avoids overdriving near switching thresholds.

The simulations abstracted away several device and circuit nonidealities that could affect EqProp in hardware. Below we summarise the most relevant effects and pragmatic countermeasures.

\subsection{Unintended State Drift}

We assumed fixed memristive states, neglecting any dynamic state transitions in the devices during free and nudging phases. If node voltages exceed device thresholds or if devices are thresholdless, the devices would experience drift in their internal states, altering the distribution of effective weights across crossbar arrays~\cite{chang_mitigating_2021, gebregiorgis_dealing_2022}. This drift can be represented by augmenting the ideal current expression with $\Delta G_{ij}$, capturing the cumulative conductance deviation arising from unintended state transitions:
\begin{equation}
    I'_j = \sum_i \bigl(G_{ij} + \Delta G_{ij}\bigr)\,U_i
\end{equation}

Even minute drifts, when accumulated over many inference cycles, can significantly alter the effective weights stored in the crossbars, leading to substantial errors in VMM computations and therefore to a degradation of loss and accuracy~\cite{chang_mitigating_2021}. To mitigate such unwanted shifts in effective weights, regular retraining or robust compensation mechanisms are necessary. One method that may be applicable to EqProp is inline calibration for memristor crossbar-based computing engine (ICE)~\cite{li_ice_2014}. ICE introduces an adaptive calibration mechanism that periodically interrupts regular operation to evaluate computational error based on benchmark data, which is then used to predict the optimal recalibration interval via polynomial fitting. Given that reprogramming a memristor to a target conductance can be up to $100$ times slower than a single inference operation, this strategy is designed to maximise the duration of continuous operation between recalibrations while maintaining acceptable levels of computational fidelity~\cite{chang_mitigating_2021}.

\subsection{Sneak-Path Currents}

Unintended currents flowing parallel to the desired path through memristors in a crossbar array form an unknown parallel resistance that depends on the states of adjacent memristors~\cite{zidan_memristor-based_2013, humood_high-density_2019}. This phenomenon can have a significant impact on the reliability of crossbar arrays, especially when array sizes increase, as sneak paths become more dominant. Figure~\ref{fig:sneak-path} illustrates this issue in a simple $2 \times 2$ crossbar array. The desired current route is depicted by the solid blue line, and the dashed red line represents a sneak path, showing how current bypasses the intended route through adjacent cells, thereby resulting in unwanted parallel conduction.
\input{figures/sneak-path}

Integrating a selector or gating device in series with each memristor cell, such as a transistor, diode, or additional memristor suppresses sneak paths~\cite{zidan_memristor-based_2013, humood_high-density_2019}. However, such selector-based approaches are not suitable for EqProp, which requires a bidirectional network architecture where all memristive devices must be simultaneously accessible to establish an equilibrium state in the NRN. One potential configuration is an unfolded crossbar, where each memristor is connected to a dedicated bitline~\cite{zidan_memristor-based_2013}. Since the final summation of the output currents must still be performed unidirectionally (e.g., implemented using transimpedance and summation amplifiers), the functionality of such a crossbar architecture is constrained to the output layer when using EqProp. In addition, this design comes at the cost of reduced memory density, resulting in an increase in the required spatial area~\cite{zidan_memristor-based_2013}. Therefore, additional studies should explore alternative strategies for EqProp-compatible sneak-path suppression mitigation.

\subsection{Parasitic Line Resistances}

Line resistances, arising from the metallic interconnects (wordlines and bitlines), distort current flow and the effective conductances by introducing spatially nonuniform voltage drops that grow with interconnect length and array size~\cite{jeong_parasitic_2018, nguyen_quantization_2022}. Figure~\ref{fig:crossbar-line-resistances} illustrates how these distributed resistances, denoted as $R_\mathrm{line}$, are present along the interconnects and accumulate over distance. Outputs located farther from the input voltage sources experience larger voltage drops along the wordlines, reducing their effective input voltages and hence their currents. This effect could incorrectly bias the network towards neurons closer to the input voltage sources. Depending on the specific implementation, $R_\mathrm{line}$ typically ranges between \SI{0.1}{\ohm} and \SI{5}{\ohm}~\cite{lambertini_simulation_2025}.
\input{figures/crossbar-line-resistances}

A method proposed to address the effect of line resistances involves adjusting the target memristance values during weight programming~\cite{nguyen_quantization_2022}. In this approach, each memristor is set to a corrected value that precompensates for the voltage drops introduced by the line resistances. The estimated parasitic resistance along the signal path is subtracted from the ideal weight matrix before mapping the values to the crossbar. However, this technique assumes unidirectional signal flow and is therefore incompatible with EqProp, as the target values would differ when the direction is reversed and the roles of inputs and outputs are exchanged.

An alternative strategy compensates for the degradation of output currents by applying predefined scaling factors at neuron level~\cite{jeong_parasitic_2018}. This method is effective because wordline resistances induce position-dependent voltage drops across the outputs, whereas bitline resistances primarily result in a uniform reduction in current magnitude. Since the required scaling factors solely depend on array size, device characteristics, and learning rules, they remain constant during network operation and can be directly implemented in hardware. A possible way to integrate this approach into EqProp could involve assigning each neuron individual scaling factors for the forward and reverse directions. For each neuron $n$, the forward scaling factor might be adjusted from $A$ to $A + \Delta^\mathrm{f}_n$, based on the properties of the preceding crossbar, while the reverse scaling factor for the subsequent layer could be modified from $\frac{1}{A}$ to $\frac{1}{A} + \Delta^\mathrm{r}_n$. Here, $\Delta^\mathrm{f}_n$ and $\Delta^\mathrm{r}_n$ represent the neuron-specific correction terms in forward and reverse directions, respectively. Whether such a compensation scheme is compatible with the dynamics of EqProp remains an open question and is left for future investigation.

\section{Conclusion}

This study investigated the effect of the nonlinear, device-specific characteristics of memristors on the convergence dynamics of loss in neural networks trained via EqProp. We found that EqProp-based learning remains robust under nonlinear, memristor-driven updates, provided that the devices offer a sufficiently large dynamic resistance range. A ratio of $\frac{R_\mathrm{OFF}}{R_\mathrm{ON}} \gtrsim 10$ is essential to enable stable analogue learning, offering a concrete fabrication target. PWM and PAM produced equivalent results for continuous-drift models. Thresholded devices required higher pulse frequencies with corresponding smaller pulse widths for PAM to avoid overshoot.

Several simplifying assumptions constrain generalisability. Memristive states were held fixed, omitting unintended drift that can accumulate during operation. Ideal crossbars were assumed, neglecting sneak-path currents and line resistances that distort bidirectional signal flow in larger arrays. SPICE runtime limited experiments to shallow networks and small datasets. Finally, device-to-device variability was not modelled explicitly. The next essential step is to conduct hardware-in-the-loop experiments using physical memristors, in order to lay the groundwork for efficient device-level implementations.

\section*{Notes}

\begin{enumerate}
    \item Source code: \url{https://github.com/Mitarano/ebana-with-memristors}
    \item An interactive memristor current-voltage simulator is available at \url{https://mitarano.github.io/memristor-simulator}.
\end{enumerate}

\section*{Disclosure Statement}

No potential conflict of interest was reported by the authors.

\bibliographystyle{tfnlm}
\bibliography{bibliography}

\end{document}

%% file: figures/crossbar-array.tex
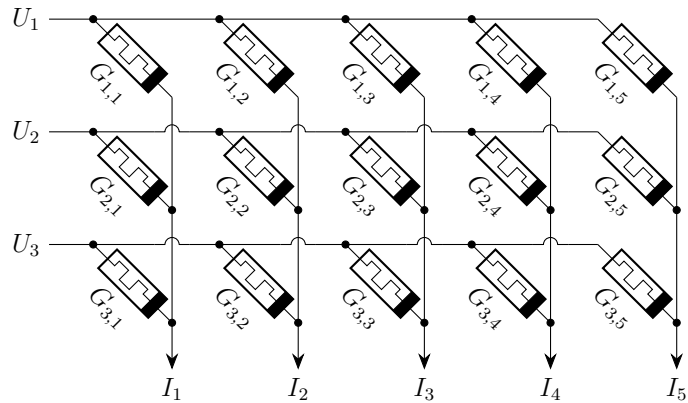
\begin{figure}[!b]
    \centering
    \begin{circuitikz}[scale=0.83, transform shape, font=\large]
        \def\cols{5}
        \def\rows{3}
        \def\colsep{2}
        \def\rowsep{1.8}
        \def\memleft{1.25}
        \def\memdown{1.25}

        \foreach \row in {1,...,\rows} {
            \pgfmathsetmacro{\ystart}{(\rows*\rowsep)-((\row-1)*\rowsep)-1}
            \draw ({-(\colsep/2)-0.2},\ystart) -- ({(1-\memleft)*\colsep},\ystart);
            \node[left] at ({-(\colsep/2)-0.2},\ystart) {$U_{\row}$};
        }
        
        \foreach \row in {1,...,\rows} {
            \foreach \col in {1,...,\cols} {
                \pgfmathsetmacro{\xstart}{(\col-\memleft)*\colsep}
                \pgfmathsetmacro{\ystart}{(\rows*\rowsep)-((\row-1)*\rowsep)-1}
                \pgfmathsetmacro{\xend}{\xstart+\memleft}
                \pgfmathsetmacro{\yend}{\ystart-\memdown}
                
                \ifnum \col=\cols
                    \ifnum \row=1
                        \draw (\xstart,\ystart) to[memristor, -, l_=$G_{\row,\col}$] (\xend,\yend);
                    \else
                        \draw (\xstart,\ystart) to[memristor, -*, l_=$G_{\row,\col}$] (\xend,\yend);
                    \fi
                \else 
                    \ifnum \row=1
                        \draw (\xstart,\ystart) to[memristor, *-, l_=$G_{\row,\col}$] (\xend,\yend);
                    \else
                        \draw (\xstart,\ystart) to[memristor, *-*, l_=$G_{\row,\col}$] (\xend,\yend);
                    \fi
                \fi
            }
        }
        
        \foreach \col in {1,...,\cols} {
            \pgfmathsetmacro{\x}{(\col-\memleft)*\colsep+\memleft}
            \pgfmathsetmacro{\y}{\rowsep-\memdown-1}
            \draw[-{Stealth[scale=1.3]}] (\x,{(\rows)*\rowsep-\memdown-1}) -- (\x,\y-\memdown*0.6);
            \node[below] at (\x,\y-\memdown*0.6) {$I_{\col}$};
        }

        \draw ({(1-\memleft)*\colsep},{\rows*\rowsep-1}) -- ({(\cols-\memleft)*\colsep},{\rows*\rowsep-1});
        
        \pgfmathsetmacro{\ymax}{(\rows-1)*\rowsep}
        \foreach \row in {2,...,\rows} {
            \foreach \col in {2,...,\cols} {
                \pgfmathsetmacro{\x}{(\col-1-\memleft)*\colsep}
                \pgfmathsetmacro{\y}{(\row-1)*\rowsep-1}
                \draw (\x,\y) -- ({\x+\memleft-0.3},\y) to[crossing, bipoles/crossing/size=0.4] ({\x+\memleft+0.3},\y) -- ({\x+\colsep},\y);
            }
        }
    \end{circuitikz}
    \caption{Basic implementation of a memristor crossbar array~\cite{aguirre_hardware_2024} for the computation of Equation~\ref{eq:vmm}.}
    \label{fig:crossbar-array}
\end{figure}

%% file: figures/line-duplication.tex
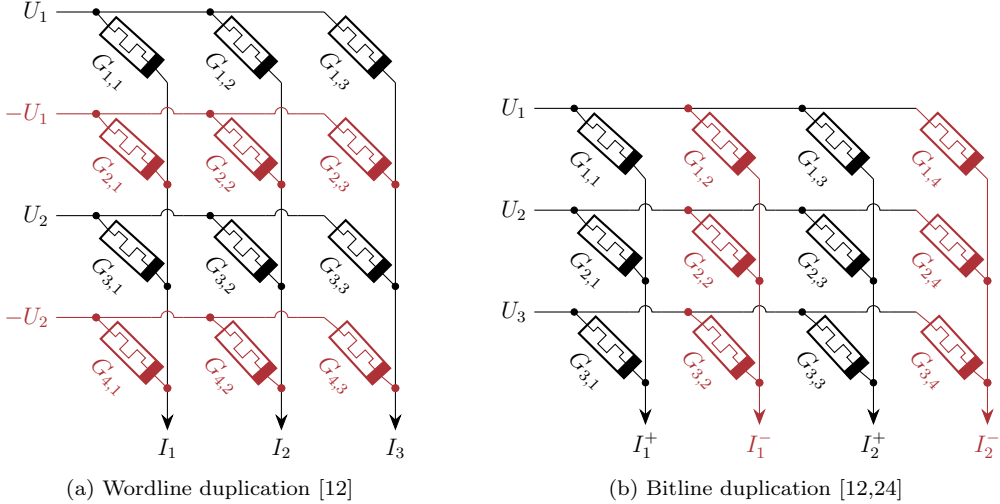
\begin{figure}[!t]
  \centering
  \subfloat[Wordline duplication~\cite{aguirre_hardware_2024}\label{fig:input-duplication}]{%
    \begin{minipage}[b]{0.44\textwidth}
      \centering 
      \begin{circuitikz}[scale=0.75, transform shape, font=\large]
        \def\cols{3}
        \def\rows{2}
        \def\colsep{2}
        \def\rowsep{1.8}
        \def\memleft{1.25}
        \def\memdown{1.25}

        \pgfmathsetmacro{\inputs}{\rows*2}
        
        \foreach \row in {1,...,\inputs} {
          \pgfmathsetmacro{\ystart}{(\inputs*\rowsep)-((\row-1)*\rowsep)-1}
          \pgfmathsetmacro{\index}{int(ceil(\row/2))}
          \pgfmathsetmacro{\modresult}{int(mod(\row,2))}
          
          \ifnum \modresult=1
              \draw ({-(\colsep/2)-0.2},\ystart) -- ({(1-\memleft)*\colsep},\ystart);
              \node[left] at ({-(\colsep/2)-0.2},\ystart) {$U_{\index}$};
          \else
              \draw[Maroon] ({-(\colsep/2)-0.2},\ystart) -- ({(1-\memleft)*\colsep},\ystart);
              \node[left, Maroon] at ({-(\colsep/2)-0.2},\ystart) {$-U_{\index}$};
          \fi
        }
        
        \foreach \col in {1,...,\cols} {
          \pgfmathsetmacro{\x}{(\col-\memleft)*\colsep+\memleft}
          \pgfmathsetmacro{\y}{\rowsep-\memdown-1}
          \draw[-{Stealth[scale=1.3]}] (\x,{(\inputs)*\rowsep-\memdown-1}) -- (\x,\y-\memdown*0.6);
          \node[below] at (\x,\y-\memdown*0.6) {$I_{\col}$};
        }
        
        \foreach \row in {1,...,\inputs} {
          \foreach \col in {1,...,\cols} {
            \pgfmathsetmacro{\xstart}{(\col-\memleft)*\colsep}
            \pgfmathsetmacro{\ystart}{(\inputs*\rowsep)-((\row-1)*\rowsep)-1}
            \pgfmathsetmacro{\xend}{\xstart+\memleft}
            \pgfmathsetmacro{\yend}{\ystart-\memdown}
            \pgfmathsetmacro{\modresult}{int(mod(\row,2))}
            
            \ifnum \col=\cols
              \ifnum \row=1
                \draw (\xstart,\ystart) to[memristor, -, l_=$G_{\row,\col}$] (\xend,\yend);
              \else
                \ifnum \modresult=1
                  \draw (\xstart,\ystart) to[memristor, -*, l_=$G_{\row,\col}$] (\xend,\yend);
                \else
                  \draw[Maroon] (\xstart,\ystart) to[memristor, -*, l_=$G_{\row,\col}$] (\xend,\yend);
                  \fill[Maroon] (\xend,\yend) circle (2pt);
                \fi
              \fi
            \else 
              \ifnum \row=1
                \draw (\xstart,\ystart) to[memristor, *-, l_=$G_{\row,\col}$] (\xend,\yend);
              \else
                \ifnum \modresult=1
                  \draw (\xstart,\ystart) to[memristor, *-*, l_=$G_{\row,\col}$] (\xend,\yend);
                \else
                  \draw[Maroon] (\xstart,\ystart) to[memristor, *-*, l_=$G_{\row,\col}$] (\xend,\yend);
                  \fill[Maroon] (\xstart,\ystart) circle (2pt);
                  \fill[Maroon] (\xend,\yend) circle (2pt);
                \fi
              \fi
            \fi
          }
        }

        \draw ({(1-\memleft)*\colsep},{\inputs*\rowsep-1}) -- ({(\cols-\memleft)*\colsep},{\inputs*\rowsep-1});
        
        \pgfmathsetmacro{\ymax}{(\inputs-1)*\rowsep}
        \foreach \row in {2,...,\inputs} {
          \foreach \col in {2,...,\cols} {
            \pgfmathsetmacro{\x}{(\col-1-\memleft)*\colsep}
            \pgfmathsetmacro{\y}{(\row-1)*\rowsep-1}
            \pgfmathsetmacro{\modresult}{int(mod(\row,2))}
            \ifnum \modresult=1
              \draw (\x,\y) -- ({\x+\memleft-0.3},\y) to[crossing, bipoles/crossing/size=0.4] ({\x+\memleft+0.3},\y) -- ({\x+\colsep},\y);
            \else
              \draw[Maroon] (\x,\y) -- ({\x+\memleft-0.3},\y) to[crossing, bipoles/crossing/size=0.4] ({\x+\memleft+0.3},\y) -- ({\x+\colsep},\y);
            \fi
          }
        }
      \end{circuitikz}
    \end{minipage}
  }
  \hfill
  \subfloat[Bitline duplication~\cite{hasan_-chip_2017, aguirre_hardware_2024}\label{fig:output-duplication}]{%
    \begin{minipage}[b]{0.54\textwidth}
      \centering
      \begin{circuitikz}[scale=0.75, transform shape, font=\large]
        \def\cols{2}
        \def\rows{3}
        \def\colsep{2}
        \def\rowsep{1.8}
        \def\memleft{1.25}
        \def\memdown{1.25}

        \pgfmathsetmacro{\outputs}{\cols*2}
        
        \foreach \row in {1,...,\rows} {
          \pgfmathsetmacro{\ystart}{(\rows*\rowsep)-((\row-1)*\rowsep)-1}
          \draw ({-(\colsep/2)-0.2},\ystart) -- ({(1-\memleft)*\colsep},\ystart);
          \node[left] at ({-(\colsep/2)-0.2},\ystart) {$U_{\row}$};
        }

        \draw ({(1-\memleft)*\colsep},{\rows*\rowsep-1}) -- ({(\outputs-\memleft)*\colsep},{\rows*\rowsep-1});
        
        \pgfmathsetmacro{\ymax}{(\rows-1)*\rowsep}
        \foreach \row in {2,...,\rows} {
          \foreach \col in {2,...,\outputs} {
            \pgfmathsetmacro{\x}{(\col-1-\memleft)*\colsep}
            \pgfmathsetmacro{\y}{(\row-1)*\rowsep-1}
            \draw (\x,\y) -- ({\x+\memleft-0.3},\y) to[crossing, bipoles/crossing/size=0.4] ({\x+\memleft+0.3},\y) -- ({\x+\colsep},\y);
          }
        }
        
        \foreach \row in {1,...,\rows} {
          \foreach \col in {1,...,\outputs} {
            \pgfmathsetmacro{\xstart}{(\col-\memleft)*\colsep}
            \pgfmathsetmacro{\ystart}{(\rows*\rowsep)-((\row-1)*\rowsep)-1}
            \pgfmathsetmacro{\xend}{\xstart+\memleft}
            \pgfmathsetmacro{\yend}{\ystart-\memdown}
            \pgfmathsetmacro{\modresult}{int(mod(\col,2))}
            
            \ifnum \col=\outputs
              \ifnum \row=1
                \ifnum \modresult=1
                  \draw (\xstart,\ystart) to[memristor, -, l_=$G_{\row,\col}$] (\xend,\yend);
                \else
                  \draw[Maroon] (\xstart,\ystart) to[memristor, -, l_=$G_{\row,\col}$] (\xend,\yend);
                \fi
              \else
                \ifnum \modresult=1
                  \draw (\xstart,\ystart) to[memristor, -*, l_=$G_{\row,\col}$] (\xend,\yend);
                \else
                  \draw[Maroon] (\xstart,\ystart) to[memristor, -*, l_=$G_{\row,\col}$] (\xend,\yend);
                  \fill[Maroon] (\xend,\yend) circle (2pt);
                \fi
              \fi
            \else 
              \ifnum \row=1
                \ifnum \modresult=1
                  \draw (\xstart,\ystart) to[memristor, *-, l_=$G_{\row,\col}$] (\xend,\yend);
                \else
                  \draw[Maroon] (\xstart,\ystart) to[memristor, *-, l_=$G_{\row,\col}$] (\xend,\yend);
                  \fill[Maroon] (\xstart,\ystart) circle (2pt);
                \fi
              \else
                \ifnum \modresult=1
                  \draw (\xstart,\ystart) to[memristor, *-*, l_=$G_{\row,\col}$] (\xend,\yend);
                \else
                  \draw[Maroon] (\xstart,\ystart) to[memristor, *-*, l_=$G_{\row,\col}$] (\xend,\yend);
                  \fill[Maroon] (\xstart,\ystart) circle (2pt);
                  \fill[Maroon] (\xend,\yend) circle (2pt);
                \fi
              \fi
            \fi
          }
        }
        
        \foreach \col in {1,...,\outputs} {
          \pgfmathsetmacro{\x}{(\col-\memleft)*\colsep+\memleft}
          \pgfmathsetmacro{\y}{\rowsep-\memdown-1}            
          \pgfmathsetmacro{\index}{int(ceil(\col/2))}
          \pgfmathsetmacro{\modresult}{int(mod(\col,2))}
          
          \ifnum \modresult=1
            \draw[-{Stealth[scale=1.3]}] (\x,{(\rows)*\rowsep-\memdown-1}) -- (\x,\y-\memdown*0.6);
            \node[below] at (\x,\y-\memdown*0.6) {$I^+_{\index}$};
          \else
            \draw[-{Stealth[scale=1.3]}, Maroon] (\x,{(\rows)*\rowsep-\memdown-1}) -- (\x,\y-\memdown*0.6);
            \node[below, Maroon] at (\x,\y-\memdown*0.6) {$I^-_{\index}$};
          \fi
        }
      \end{circuitikz}
    \end{minipage}
  }
  \caption{Negative-weight emulation in memristor crossbars using (a) wordline duplication and (b) bitline duplication. Additional lines and memristors are highlighted in red.}
  \label{fig:line-duplication}
\end{figure}

%% file: figures/eqprop-flowchart.tex
\begin{figure}[!b]
    \centering
    \tikzstyle{decision} = [diamond, draw, fill=Maroon!20, 
    text width=6.5em, text badly centered, node distance=3cm, inner sep=-3pt, minimum width=8em, minimum height=8em,]
    \tikzstyle{block} = [rectangle, draw, fill=RoyalBlue!20, 
        text width=9.5em, text centered, rounded corners, minimum height=4em, node distance = 2.25cm]
    \tikzstyle{line} = [draw, -{Stealth[scale=1.2]}]
    \tikzstyle{cloud} = [draw, ellipse,fill=red!20, node distance=4.5cm,
        minimum height=2em, text width=6.5em, text centered]
    \begin{tikzpicture}[scale=0.73, transform shape, auto]
        \node [block] (fix) {fix $x$ at\\ input nodes};
        \node [block, below of=fix] (free) {settle to $s_0$};
        \node [block, below of=free, node distance=3cm] (nudged) {nudge output nodes towards their target and settle to $s_\beta$};
        \node [cloud, right of=free, fill=Maroon!20, node distance=4.6cm] (free_gradient) {collect $\frac{\partial E}{\partial \theta}(\theta, x, s_0)$};
        \node [cloud, right of=nudged, fill=Maroon!20, node distance=4.6cm] (nudged_gradient) {collect $\frac{\partial E}{\partial \theta}(\theta, x, s_\beta)$};
        \node [block, right of=free_gradient, node distance=4.8cm] (update) {calculate $\frac{\partial \mathcal{L}_{\beta}}{\partial \theta}(\theta, x, y)$ and update $\theta$};
        
        \path [line] (fix) -- (free);
        \path [line] (free) -- (nudged);
        \path [line,dashed] (free) -- (free_gradient);
        \path [line,dashed] (nudged) -- (nudged_gradient);
        \path [line] (free_gradient) -- (update);
        \path [line] (nudged_gradient) -| (update);
        \path [line] (update) |- node[pos=0.75, above]{next sample} (fix);
        
        \begin{pgfonlayer}{background}
            \node[inner sep=0pt] (free_top) at ([yshift=-12pt]free.south) {};
            \node[fill=gray!16, rounded corners=7pt, draw=none, 
                  fit=(free_top) (free) (free_gradient), inner sep=8pt] (freebox) {};
            \node[anchor=east, align=right] 
                at ([xshift=-5pt,yshift=8pt]freebox.south east) {\textbf{Free Phase}};
        
            \node[inner sep=0pt] (nudged_top) at ([yshift=-12pt]nudged.south) {};
            \node[fill=gray!16, rounded corners=7pt, draw=none, 
                  fit=(nudged_top) (nudged) (nudged_gradient), inner sep=8pt] (nudgedbox) {};
            \node[anchor=east, align=right] 
                at ([xshift=-5pt,yshift=8pt]nudgedbox.south east) {\textbf{Nudging Phase}};
        \end{pgfonlayer}
    \end{tikzpicture}
    \caption{Flowchart of the EqProp approach~\cite{scellierEquilibriumPropagationBridging2017,watfaEnergybasedAnalogNeural2023} with stochastic gradient descent.}
    \label{fig:eqprop-flowchart}
\end{figure}
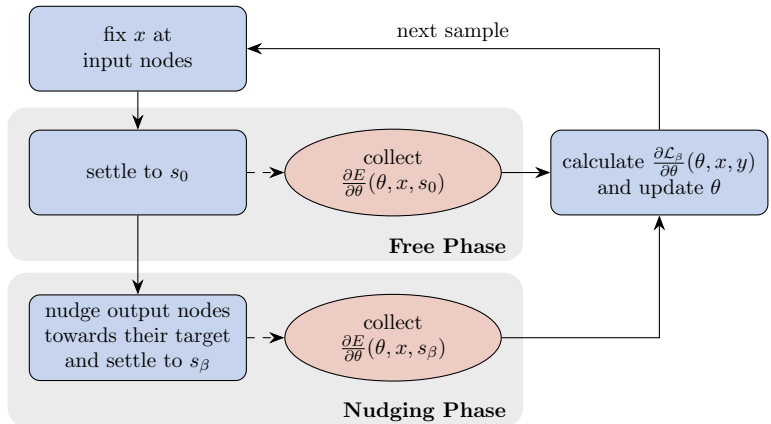

%% file: figures/non-linearity.tex
\begin{figure}[!b]
    \centering
    \subfloat[Circuit of a ReLU-like nonlinearity~\cite{watfaEnergybasedAnalogNeural2023}\label{fig:relu-circuit}]{%
    \begin{minipage}[t][2cm][t]{0.3\textwidth}
        \centering
        \begin{circuitikz}[american, scale=0.75, transform shape]
            \draw (0,3) node[circle, draw, fill=white, inner sep=0.5mm] (input) {};
            \draw (3.5,3) node[circle, draw, fill=white, inner sep=0.5mm] (output) {};
            \draw (1.75,3) -- (1.75,2.5) to[diode] (1.75,1.4) to[voltage source] (1.75,0) -- node[ground]{} (1.75,0);
            \draw (input) -- (output);
        \end{circuitikz}
    \end{minipage}
    }
    \hfill
    \subfloat[Circuit of a sigmoid-like nonlinearity~\cite{kendall_training_2020}\label{fig:sigmoid-circuit}]{%
    \begin{minipage}[t][2cm][t]{0.3\textwidth}
        \centering
        \begin{circuitikz}[american, scale=0.75, transform shape]
            \draw (0,3) node[circle, draw, fill=white, inner sep=0.5mm] (input) {};
            \draw (3.5,3) node[circle, draw, fill=white, inner sep=0.5mm] (output) {};
            \draw (1.75,3) -- (1.75,2.5) -- (1,2.5) to[diode] (1,1.4) to[voltage source] (1,0) -- (1.75,0) -- node[ground]{} (1.75,0);
            \draw (1.75,2.5) -- (2.5,2.5) to[diode, invert] (2.5,1.4) to[voltage source, invert] (2.5,0) -- (1.75,0);
            \draw (input) -- (output);
        \end{circuitikz}
    \end{minipage}
    }
    \hfill
    \subfloat[Bidirectional amplifier~\cite{kendall_training_2020}\label{fig:bidirectional-amplifier}]{%
    \begin{minipage}[t]{0.3\textwidth}
        \begin{circuitikz}[american, scale=0.75, transform shape]
            \draw (-0.5,3) node[circle, draw, fill=white, inner sep=0.5mm] (input) {};
            \draw (4.8,3) node[circle, draw, fill=white, inner sep=0.5mm] (output) {};
            \draw (2.435,2) node[circle, draw, fill=white, inner sep=0.5mm] (voltage) {};
            \draw (1.575,2) node[circle, draw, fill=white, inner sep=0.5mm] (current) {};
            \draw (input) -- (1,3) to[controlled current source] node[ground]{} (1,1);
            \draw (output) -- (3,3) to[controlled voltage source] node[ground]{} (3,1);
            \draw[dashed] (1,3) -- (2.15,3) -- (2.15,2) -- (voltage);
            \draw[densely dashed] (3.5,2) -- (3.8,2) -- (3.8,0.25) -- (1.85,0.25) -- (1.85,2) -- (current);
        \end{circuitikz}
    \end{minipage}
    }
    
    \subfloat[Voltage-current characteristic of the ReLU-like nonlinearity\label{fig:relu}]{%
    \begin{minipage}[t]{0.3\textwidth}
        \centering
        \begin{tikzpicture}
            \begin{axis}[
                height=4.5cm,
                width=5.5cm,
                grid=both,
                grid style={line width=.1pt, draw=gray!10},
                major grid style={line width=.2pt,draw=gray!50},
                axis lines=middle,
                axis line style={-stealth},
                enlargelimits=true,
                xlabel={$I$},
                xlabel style={
                    at={(ticklabel* cs:1)},
                    anchor=north
                },
                ylabel={$U$},
                ylabel style={
                    at={(ticklabel* cs:1)},
                    anchor=east
                },
                xtick=\empty,
                ytick=\empty,
                xmin=-2.5, xmax=2.5,
                ymin=-1.7, ymax=1.7,
            ]
                \addplot[
                    color=RoyalBlue,
                    ultra thick,
                ] table[
                    col sep=comma,
                    x=I_in,
                    y=V_out
                ] {data/relu.csv};
            \end{axis}
        \end{tikzpicture}
    \end{minipage}
    }
    \hfill
    \subfloat[Voltage-current characteristic of the sigmoid-like nonlinearity\label{fig:sigmoid}]{%
    \begin{minipage}[t]{0.3\textwidth}
        \centering
        \begin{tikzpicture}
            \begin{axis}[
                height=4.5cm,
                width=5.5cm,
                grid=both,
                grid style={line width=.1pt, draw=gray!10},
                major grid style={line width=.2pt,draw=gray!50},
                axis lines=middle,
                axis line style={-stealth},
                enlargelimits=true,
                xlabel={$I$},
                xlabel style={
                    at={(ticklabel* cs:1)},
                    anchor=north
                },
                ylabel={$U$},
                ylabel style={
                    at={(ticklabel* cs:1)},
                    anchor=east
                },
                xtick=\empty,
                ytick=\empty,
                xmin=-2.5, xmax=2.5,
                ymin=-1.7, ymax=1.7,
            ]
                \addplot[
                    color=RoyalBlue,
                    ultra thick,
                ] table[
                    col sep=comma,
                    x=I_in,
                    y=V_out
                ] {data/sigmoid.csv};
            \end{axis}
        \end{tikzpicture}
    \end{minipage}
    }
    \hfill
    \makebox[0.3\textwidth]{}
    \caption{Circuits for the implementation of an analog neuron (top) and the corresponding voltage-current characteristics of the nonlinearities (bottom).}
    \label{fig:neuron-circuits}
\end{figure}
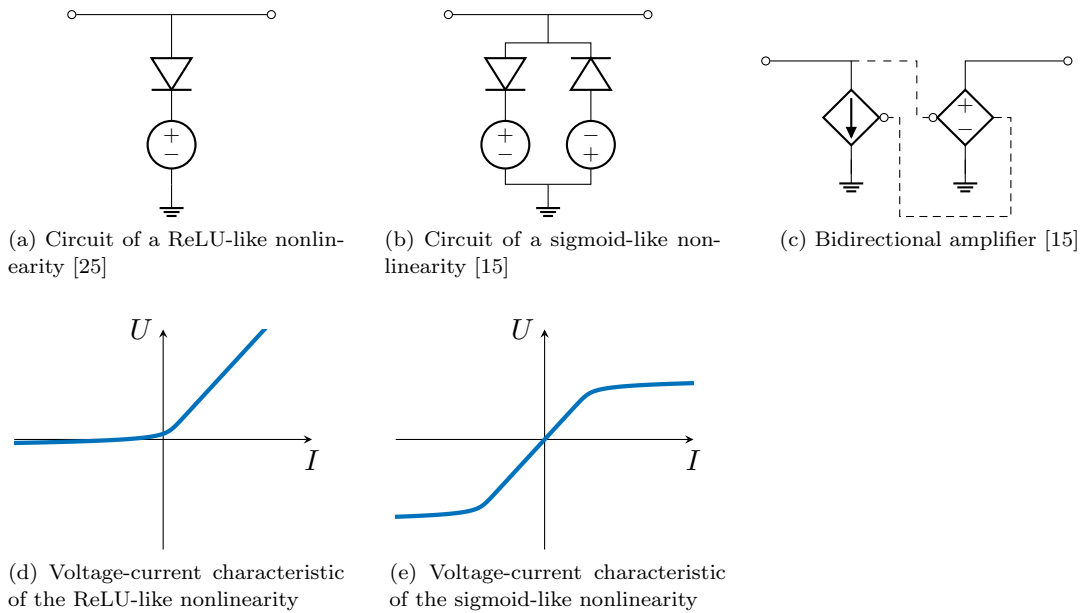

%% file: figures/nrn-architecture.tex
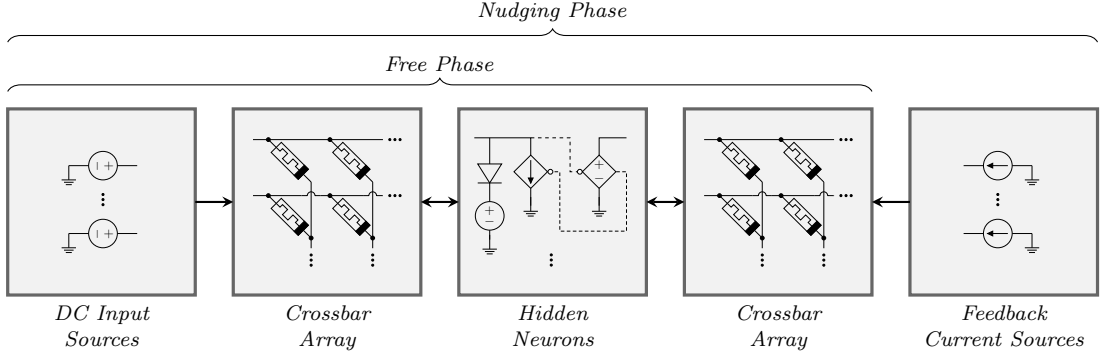
\begin{figure}
\centering

\newlength{\gspace}
\setlength{\gspace}{5mm} 

\newlength{\sqsize}
\setlength{\sqsize}{\dimexpr(\linewidth - 4\gspace)/5}

\newcommand{\mylabel}[1]{%
    \footnotesize\textit{\parbox{\dimexpr(\sqsize-3mm)}{\centering #1}}
}

\begin{tikzpicture}[american,>=stealth]

\draw[draw=black!60, fill=black!5, very thick] (0,0) rectangle (\sqsize,\sqsize);
\node[below] at (\sqsize/2,0) {\mylabel{DC Input \\ Sources}};
\coordinate (R1) at (\sqsize,\sqsize/2);

\node[anchor=center] at (\sqsize/2,\sqsize/2) {%
    \scalebox{0.45}{%
        \begin{circuitikz}
            \draw (0,2) node[ground]{} to[voltage source, invert] (2,2);
            \draw[fill=black] (1,1.15) circle (0.03);
            \draw[fill=black] (1,1) circle (0.03);
            \draw[fill=black] (1,0.85) circle (0.03);
            \draw (0,0) node[ground]{} to[voltage source, invert] (2,0);
        \end{circuitikz}
    }%
};

\draw[draw=black!60, fill=black!5, very thick] (\sqsize+\gspace,0) rectangle (\sqsize+\gspace+\sqsize,\sqsize);
\node[below] at (\sqsize+\gspace+\sqsize/2,0) {\mylabel{Crossbar \\ Array}};
\coordinate (L2) at (\sqsize+\gspace,\sqsize/2);
\coordinate (R2) at (\sqsize+\gspace+\sqsize,\sqsize/2);

\node[anchor=center] at (\sqsize+\gspace+\sqsize/2,\sqsize/2) {%
    \scalebox{0.45}{%
        \begin{circuitikz}[clip=false]
            \def\cols{2}
            \def\rows{2}
            \def\colsep{1.8}
            \def\rowsep{1.65}
            \def\memleft{1.25}
            \def\memdown{1.25}

            \foreach \row in {1,...,\rows} {
                \pgfmathsetmacro{\ystart}{(\rows*\rowsep)-((\row-1)*\rowsep)-1}
                \draw ({-\colsep/2},\ystart) -- ({(1-\memleft)*\colsep},\ystart);
            }

            \foreach \row in {1,...,\rows} {
                \foreach \col in {1,...,\cols} {
                    \pgfmathsetmacro{\xstart}{(\col-\memleft)*\colsep}
                    \pgfmathsetmacro{\ystart}{(\rows*\rowsep)-((\row-1)*\rowsep)-1}
                    \pgfmathsetmacro{\xend}{\xstart+\memleft}
                    \pgfmathsetmacro{\yend}{\ystart-\memdown}
                    
                    \ifnum \row=1
                        \draw (\xstart,\ystart) to[memristor, *-] (\xend,\yend);
                    \else
                        \draw (\xstart,\ystart) to[memristor, *-*] (\xend,\yend);
                    \fi
                }
            }

            \foreach \col in {1,...,\cols} {
                \pgfmathsetmacro{\x}{(\col-\memleft)*\colsep+\memleft}
                \pgfmathsetmacro{\y}{\rowsep-\memdown-1}
                \draw (\x,{(\rows)*\rowsep-\memdown-1}) -- (\x,{\y-0.3});
                \draw[fill=black] (\x,{\y-0.35-0.15}) circle (0.03);
                \draw[fill=black] (\x,{\y-0.35-0.3}) circle (0.03);
                \draw[fill=black] (\x,{\y-0.35-0.45}) circle (0.03);
            }

            \draw ({(1-\memleft)*\colsep},{\rows*\rowsep-1}) -- ({(\cols-\memleft)*\colsep+\memleft+0.3},{\rows*\rowsep-1});
            \draw[fill=black] ({(\cols-\memleft)*\colsep+\memleft+0.35+0.15},{\rows*\rowsep-1}) circle (0.03);
            \draw[fill=black] ({(\cols-\memleft)*\colsep+\memleft+0.35+0.3},{\rows*\rowsep-1}) circle (0.03);
            \draw[fill=black] ({(\cols-\memleft)*\colsep+\memleft+0.35+0.45},{\rows*\rowsep-1}) circle (0.03);

            \pgfmathsetmacro{\ymax}{(\rows-1)*\rowsep}
            \foreach \row in {2,...,\rows} {
                \foreach \col in {1,...,\cols} {
                    \pgfmathsetmacro{\x}{(\col-\memleft)*\colsep}
                    \pgfmathsetmacro{\y}{(\row-1)*\rowsep-1}
                    \ifnum \col=\cols
                        \draw (\x,\y) -- ({\x+\memleft-0.3},\y) to[crossing, bipoles/crossing/size=0.4] ({\x+\memleft+0.3},\y);
                        \draw[fill=black] ({\x+\memleft+0.35+0.15},\y) circle (0.03);
                        \draw[fill=black] ({\x+\memleft+0.35+0.3},\y) circle (0.03);
                        \draw[fill=black] ({\x+\memleft+0.35+0.45},\y) circle (0.03);
                    \else
                        \draw (\x,\y) -- ({\x+\memleft-0.3},\y) to[crossing, bipoles/crossing/size=0.4] ({\x+\memleft+0.3},\y) -- ({\x+\colsep},\y);
                    \fi
                }
            }
        \end{circuitikz}%
    }%
};

\draw[draw=black!60, fill=black!5, very thick] (2*\sqsize+2*\gspace,0) rectangle (2*\sqsize+2*\gspace+\sqsize,\sqsize);
\node[below] at (2*\sqsize+2*\gspace+\sqsize/2,0) {\mylabel{Hidden \\ Neurons}};
\coordinate (L3) at (2*\sqsize+2*\gspace,\sqsize/2);
\coordinate (R3) at (2*\sqsize+2*\gspace+\sqsize,\sqsize/2);

\node[anchor=center] at (2*\sqsize+2*\gspace+\sqsize/2,\sqsize/2) {%
    \scalebox{0.45}{
        \begin{circuitikz}[>=stealth]
            \draw (2.435,2) node[circle, draw, fill=white, inner sep=0.5mm] (voltage) {};
            \draw (1.575,2) node[circle, draw, fill=white, inner sep=0.5mm] (current) {};
            \draw (-0.2,3) -- (-0.2,2.5) to[diode] (-0.2,1.4) to[voltage source] node[ground]{} (-0.2,0);
            \draw (-0.65,3) -- (1,3) to[controlled current source] node[ground]{} (1,1);
            \draw (3.8,3) -- (3,3) to[controlled voltage source] node[ground]{} (3,1);
            \draw[dashed] (1,3) -- (2.15,3) -- (2.15,2) -- (voltage);
            \draw[densely dashed] (3.5,2) -- (3.8,2) -- (3.8,0.25) -- (1.85,0.25) -- (1.85,2) -- (current);
            
            \draw[fill=black] (1.6,-0.45) circle (0.03);
            \draw[fill=black] (1.6,-0.6) circle (0.03);
            \draw[fill=black] (1.6,-0.75) circle (0.03);
        \end{circuitikz}
    }%
};

\draw[draw=black!60, fill=black!5, very thick] (3*\sqsize+3*\gspace,0) rectangle (3*\sqsize+3*\gspace+\sqsize,\sqsize);
\node[below] at (3*\sqsize+3*\gspace+\sqsize/2,0) {\mylabel{Crossbar \\ Array}};
\coordinate (L4) at (3*\sqsize+3*\gspace,\sqsize/2);
\coordinate (R4) at (3*\sqsize+3*\gspace+\sqsize,\sqsize/2);

\node[anchor=center] at (3*\sqsize+3*\gspace+\sqsize/2,\sqsize/2) {%
    \scalebox{0.45}{%
        \begin{circuitikz}
            \def\cols{2}
            \def\rows{2}
            \def\colsep{1.8}
            \def\rowsep{1.65}
            \def\memleft{1.25}
            \def\memdown{1.25}

            \foreach \row in {1,...,\rows} {
                \pgfmathsetmacro{\ystart}{(\rows*\rowsep)-((\row-1)*\rowsep)-1}
                \draw ({-\colsep/2},\ystart) -- ({(1-\memleft)*\colsep},\ystart);
            }

            \foreach \row in {1,...,\rows} {
                \foreach \col in {1,...,\cols} {
                    \pgfmathsetmacro{\xstart}{(\col-\memleft)*\colsep}
                    \pgfmathsetmacro{\ystart}{(\rows*\rowsep)-((\row-1)*\rowsep)-1}
                    \pgfmathsetmacro{\xend}{\xstart+\memleft}
                    \pgfmathsetmacro{\yend}{\ystart-\memdown}
                    
                    \ifnum \row=1
                        \draw (\xstart,\ystart) to[memristor, *-] (\xend,\yend);
                    \else
                        \draw (\xstart,\ystart) to[memristor, *-*] (\xend,\yend);
                    \fi
                }
            }

            \foreach \col in {1,...,\cols} {
                \pgfmathsetmacro{\x}{(\col-\memleft)*\colsep+\memleft}
                \pgfmathsetmacro{\y}{\rowsep-\memdown-1}
                \draw (\x,{(\rows)*\rowsep-\memdown-1}) -- (\x,{\y-0.3});
                \draw[fill=black] (\x,{\y-0.35-0.15}) circle (0.03);
                \draw[fill=black] (\x,{\y-0.35-0.3}) circle (0.03);
                \draw[fill=black] (\x,{\y-0.35-0.45}) circle (0.03);
            }

            \draw ({(1-\memleft)*\colsep},{\rows*\rowsep-1}) -- ({(\cols-\memleft)*\colsep+\memleft+0.3},{\rows*\rowsep-1});
            \draw[fill=black] ({(\cols-\memleft)*\colsep+\memleft+0.35+0.15},{\rows*\rowsep-1}) circle (0.03);
            \draw[fill=black] ({(\cols-\memleft)*\colsep+\memleft+0.35+0.3},{\rows*\rowsep-1}) circle (0.03);
            \draw[fill=black] ({(\cols-\memleft)*\colsep+\memleft+0.35+0.45},{\rows*\rowsep-1}) circle (0.03);

            \pgfmathsetmacro{\ymax}{(\rows-1)*\rowsep}
            \foreach \row in {2,...,\rows} {
                \foreach \col in {1,...,\cols} {
                    \pgfmathsetmacro{\x}{(\col-\memleft)*\colsep}
                    \pgfmathsetmacro{\y}{(\row-1)*\rowsep-1}
                    \ifnum \col=\cols
                        \draw (\x,\y) -- ({\x+\memleft-0.3},\y) to[crossing, bipoles/crossing/size=0.4] ({\x+\memleft+0.3},\y);
                        \draw[fill=black] ({\x+\memleft+0.35+0.15},\y) circle (0.03);
                        \draw[fill=black] ({\x+\memleft+0.35+0.3},\y) circle (0.03);
                        \draw[fill=black] ({\x+\memleft+0.35+0.45},\y) circle (0.03);
                    \else
                        \draw (\x,\y) -- ({\x+\memleft-0.3},\y) to[crossing, bipoles/crossing/size=0.4] ({\x+\memleft+0.3},\y) -- ({\x+\colsep},\y);
                    \fi
                }
            }
        \end{circuitikz}%
    }%
};

\draw[draw=black!60, fill=black!5, very thick] (4*\sqsize+4*\gspace,0) rectangle (4*\sqsize+4*\gspace+\sqsize,\sqsize);
\node[below] at (4*\sqsize+4*\gspace+\sqsize/2,0) {\mylabel{Feedback \\ Current Sources}};
\coordinate (L5) at (4*\sqsize+4*\gspace,\sqsize/2);

\node[anchor=center] at (4*\sqsize+4*\gspace+\sqsize/2,\sqsize/2) {%
    \scalebox{0.45}{%
        \begin{circuitikz}
            \draw (2,2) node[ground]{} to[current source] (0,2);
            \draw[fill=black] (1,1.15) circle (0.03);
            \draw[fill=black] (1,1) circle (0.03);
            \draw[fill=black] (1,0.85) circle (0.03);
            \draw (2,0) node[ground]{} to[current source] (0,0);
        \end{circuitikz}
    }%
};

\draw[->,thick] (R1) -- (L2);
\draw[<->,thick] (R2) -- (L3);
\draw[<->,thick] (R3) -- (L4);
\draw[<-,thick] (R4) -- (L5);

\draw[decorate,decoration={brace,amplitude=5pt},yshift=0.6em]
(0,\sqsize) -- (3*\sqsize+3*\gspace+\sqsize,\sqsize) node[midway,above=4pt]{\textit{\footnotesize Free Phase}};

\draw[decorate,decoration={brace,amplitude=5pt},yshift=2.3em]
(0,\sqsize) -- (4*\sqsize+4*\gspace+\sqsize,\sqsize) node[midway,above=4pt]{\textit{\footnotesize Nudging Phase}};

\end{tikzpicture}\\[-0.5em]
\caption{Example architecture of a layered NRN using EqProp~\cite{kirazImpactsFeedbackCurrent2022}.}
\label{fig:nrn-architecture}
\end{figure}

%% file: figures/modulation-schemes.tex
\begin{figure}[!h]
    \centering
    \subfloat[PWM\label{fig:pwm}]{
    \begin{minipage}[t]{0.5\textwidth}
        \begin{tikzpicture}
            \draw[-stealth] (0.6,0) -- (5.7,0) node[midway, below] {$t$};
            \draw[-stealth] (0.6,0) -- (0.6,2.2) node[midway, above, rotate=90] {$u$};
            
            \def\pwmamp{1.4}
            
            \foreach \x/\w in {1/0.4, 2/1, 3/0.6, 4/1.2, 5/0.8} {
                \draw[fill=RoyalBlue] (\x-\w/3,0) rectangle (\x+\w/3,\pwmamp);
            }
        \end{tikzpicture}
    \end{minipage}
    }
    \subfloat[PAM\label{fig:pam}]{
    \begin{minipage}[t]{0.5\textwidth}
        \begin{tikzpicture}
            \draw[-stealth] (0.4,0) -- (5.7,0) node[midway, below] {$t$};
            \draw[-stealth] (0.4,0) -- (0.4,2.2) node[midway, above, rotate=90] {$u$};
            
            \foreach \x/\amp in {1/0.6, 2/1.5, 3/0.9, 4/1.8, 5/1.2} {
                \draw[fill=Maroon] (\x-0.3,0) rectangle (\x+0.3,\amp);
            }
        \end{tikzpicture}
    \end{minipage}
    }
    \caption[Comparison of pulse modulation schemes]{Comparison of pulse modulation schemes~\cite{tomasi_advanced_2014}.}
    \label{fig:modulation-schemes}
\end{figure}
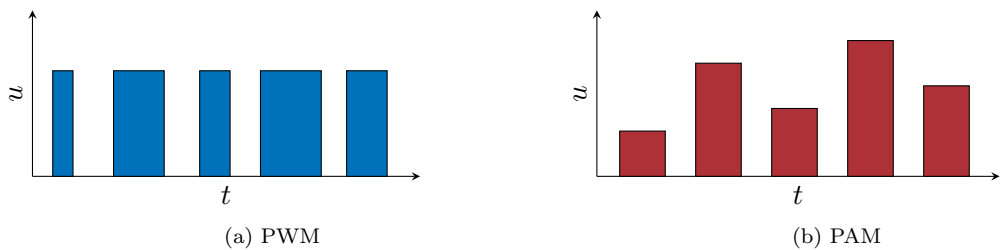

%% file: figures/modulation-updates.tex
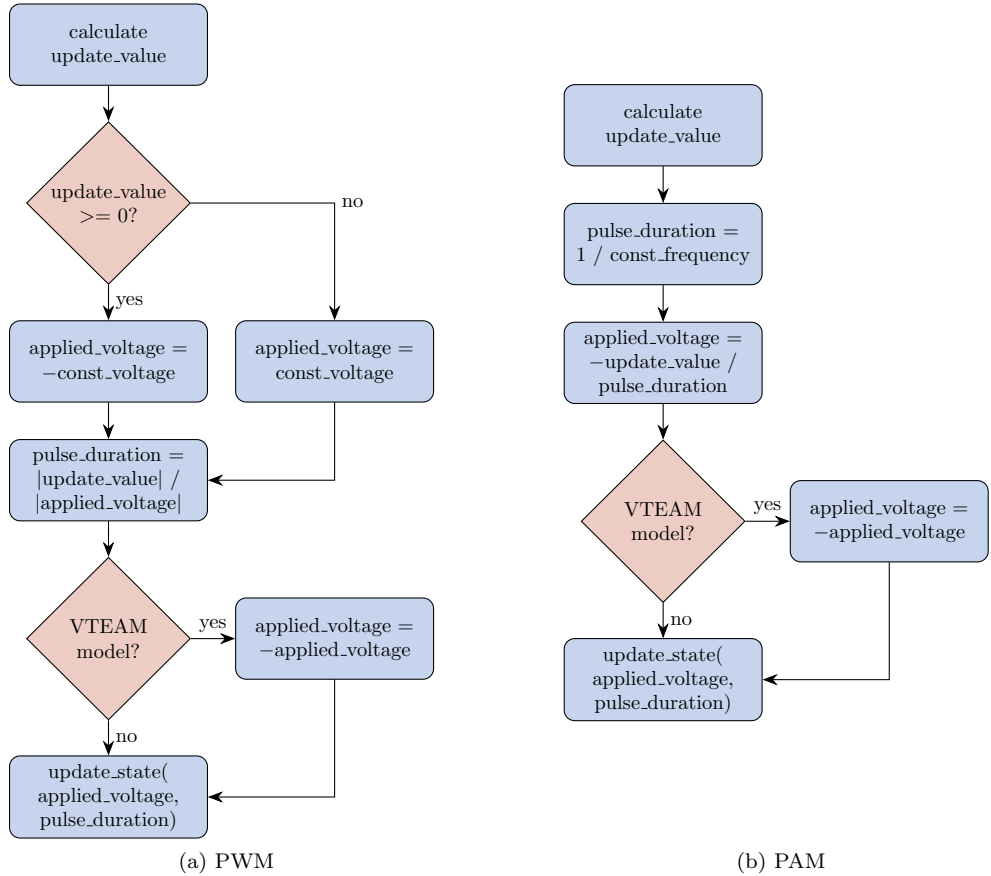
\begin{figure}[!h]
    \centering
    \tikzstyle{decision} = [diamond, draw, fill=Maroon!20, 
    text width=6em, text badly centered, node distance=3cm, inner sep=-3pt, minimum width=8em, minimum height=8em,]
    \tikzstyle{block} = [rectangle, draw, fill=RoyalBlue!20, 
        text width=9em, text centered, rounded corners, minimum height=4em, node distance = 2.25cm]
    \tikzstyle{line} = [draw, -{Stealth[scale=1.2]}]
    \subfloat[PWM\label{fig:pwm-updates}]{%
    \begin{minipage}[b]{0.5\textwidth}
        \centering
        \begin{tikzpicture}[scale=0.7, transform shape, auto]
            \node [block] (start) {calculate\\ update\_value};
            \node [decision, below of=start] (decide) {update\_value $>=$ $0$?};
            \node [block, below of=decide, node distance=3cm] (voltage1) {applied\_voltage $=$\\ $-$const\_voltage};
            \node [block, right of=voltage1, node distance=4.25cm] (voltage2) {applied\_voltage $=$\\ const\_voltage};
            \node [block, below of=voltage1] (pulse_duration) {pulse\_duration = $|\mathrm{update\_value}|$ $/$ $|\mathrm{applied\_voltage}|$};
            \node [decision, below of=pulse_duration] (decide2) {VTEAM model?};
            \node [block, right of=decide2, node distance=4.25cm] (voltage3) {applied\_voltage $=$ $-$applied\_voltage};
            \node [block, below of=decide2, node distance=3cm] (update) {update\_state( applied\_voltage, pulse\_duration)};
            
            \path [line] (start) -- (decide);
            \path [line] (decide) -- node {yes}(voltage1);
            \path [line] (decide) -| node {no}(voltage2);
            \path [line] (voltage1) -- (pulse_duration);
            \path [line] (voltage2) |- (pulse_duration);
            \path [line] (pulse_duration) -- (decide2);
            \path [line] (decide2) -- node {yes}(voltage3);
            \path [line] (decide2) -- node {no}(update);
            \path [line] (voltage3) |- (update);
        \end{tikzpicture}
    \end{minipage}
    }%
    \subfloat[PAM\label{fig:pam-updates}]{%
    \begin{minipage}[b]{0.5\textwidth}
        \centering
        \begin{tikzpicture}[scale=0.7, transform shape, auto]
            \node [block] (start) {calculate\\ update\_value};
            \node [block, below of=start] (pulse_duration) {pulse\_duration =\\ $1$ $/$ const\_frequency};
            \node [block, below of=pulse_duration] (voltage1) {applied\_voltage =\\ $-$update\_value $/$\\ pulse\_duration};
            \node [decision, below of=voltage1] (decide2) {VTEAM model?};
            \node [block, right of=decide2, node distance=4.25cm] (voltage3) {applied\_voltage $=$ $-$applied\_voltage};
            \node [block, below of=decide2, node distance=3cm] (update) {update\_state( applied\_voltage, pulse\_duration)};
            
            \path [line] (start) -- (pulse_duration);
            \path [line] (pulse_duration) -- (voltage1);
            \path [line] (voltage1) -- (decide2);
            \path [line] (decide2) -- node {yes}(voltage3);
            \path [line] (decide2) -- node {no}(update);
            \path [line] (voltage3) |- (update);
        \end{tikzpicture}
        \par\vspace{1.556cm}%
    \end{minipage}
    }%
    \caption{Flowcharts of the PWM and PAM implementations.}
    \label{fig:modulation-updates}
\end{figure}

%% file: figures/yakopcic-current.tex
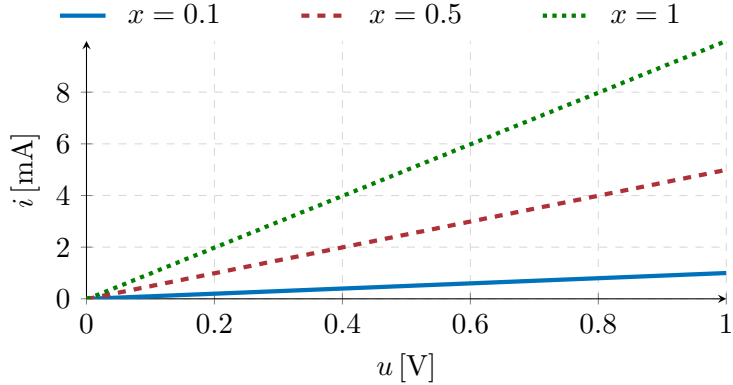
\begin{figure}[!h]
    \centering
    \begin{tikzpicture}
        \begin{axis}[
            hide axis,
            xmin=0, xmax=1,
            ymin=0, ymax=1,
            legend columns=3,
            legend style={
                draw=none,
                anchor=north,
                column sep=0.5em,
                /tikz/every even column/.append style={column sep=2.5em},
            }
        ]
        \addlegendimage{RoyalBlue, solid, ultra thick, no marks}
        \addlegendentry{$x=0.1$}
        \addlegendimage{Maroon, dashed, ultra thick, no marks}
        \addlegendentry{$x=0.5$}
        \addlegendimage{webgreen, dotted, ultra thick, no marks}
        \addlegendentry{$x=1$}
        \end{axis}
    \end{tikzpicture}
    \vspace{-5mm}
    \begin{tikzpicture}
        \begin{axis}[
            width=10cm,
            height=5cm,
            xlabel={$u \, [\si{\volt}]$},
            ylabel={$i \, [\si{\milli\ampere}]$},
            axis lines = left,
            grid=major,
            grid style={dashed,gray!30},
            xmin=0, xmax=1,
            ylabel style={yshift=-0.5mm},
        ]
        \node[anchor=west] at (axis cs:1.8, 0.2) {$a_{\text{on}}$};
        \addplot[
            RoyalBlue,
            solid,
            domain=0:1,
            samples=201,
            ultra thick,
        ]
        {1000 * 0.2 * 0.1 * sinh(0.05 * x)};
        \addplot[
            Maroon,
            dashed,
            domain=0:1,
            samples=201,
            ultra thick,
        ]
        {1000 * 0.2 * 0.5 * sinh(0.05 * x)};
        \addplot[
            webgreen,
            dotted,
            domain=0:1,
            samples=201,
            ultra thick,
        ]
        {1000 * 0.2 * 1 * sinh(0.05 * x)};
        \end{axis}
    \end{tikzpicture}
    \caption{Current-voltage relationship of the Yakopcic memristor model ($b=0.05$ and $a=0.2$).}
    \label{fig:yakopcic-current}
\end{figure}

%% file: figures/hysteresis-comparison.tex
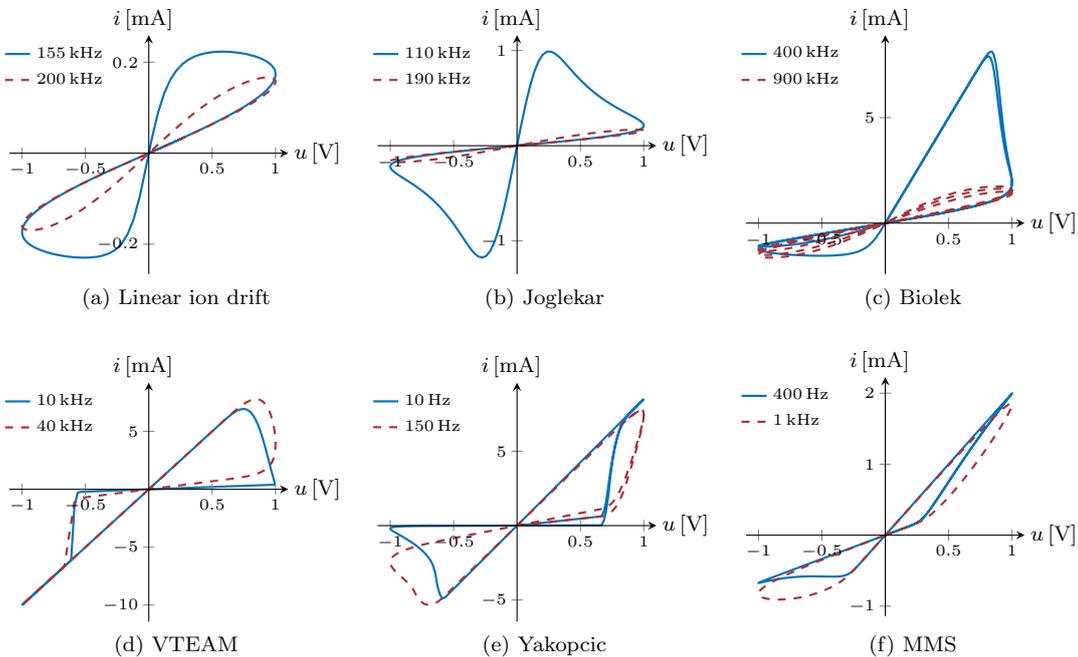
\begin{figure}[!b]
\centering
\begingroup
  \pgfplotsset{
    hyst/.style={
      height=4.75cm,
      width=5.25cm,
      axis lines=middle,
      axis line style={-stealth},
      xlabel={$u\,[\si{\volt}]$},
      xlabel style={
        font=\scriptsize,
        at={(ticklabel* cs:1)},
        anchor=north,
        xshift=11pt,
        yshift=8pt
      },
      ylabel={$i\,[\si{\milli\ampere}]$},
      ylabel style={
        font=\scriptsize,
        at={(ticklabel* cs:1)},
        anchor=east,
        xshift=13pt,
        yshift=6pt
      },
      xticklabel style={yshift=2pt},
      yticklabel style={xshift=2pt}, 
      xmin=-1.1,
      xmax=1.1,
      enlarge y limits=0.08,
      tick label style={
        /pgf/number format/fixed,
        font=\tiny,
      },
      axis on top=true,
      grid=none,
      legend style={
        font=\tiny,
        draw=none,
        fill opacity=0,
        text opacity=1,
        at={(rel axis cs:-0.05,1)},
        anchor=north west
      },
      legend cell align=left,
      legend columns=1,
      legend image post style={scale=0.55}
    },
  }%

  \subfloat[Linear ion drift]{
    \begin{minipage}[t]{0.32\textwidth}\centering\hspace*{-2mm}
      \begin{tikzpicture}
        \begin{axis}[hyst]
          \addplot[RoyalBlue, thick]
            table[col sep=comma, x=V, y expr=\thisrow{I}*1000]
            {data/hysteresis/linear_ion_drift_1.csv};
          \addlegendentry{\SI{155}{\kilo\hertz}}
          \addplot[Maroon, thick, dashed]
            table[col sep=comma, x=V, y expr=\thisrow{I}*1000]
            {data/hysteresis/linear_ion_drift_2.csv};
          \addlegendentry{\SI{200}{\kilo\hertz}}
        \end{axis}
      \end{tikzpicture}
    \end{minipage}
  }%
  \subfloat[Joglekar]{
    \begin{minipage}[t]{0.32\textwidth}\centering\hspace*{-2mm}
      \begin{tikzpicture}
        \begin{axis}[hyst]
          \addplot[RoyalBlue, thick]
            table[col sep=comma, x=V, y expr=\thisrow{I}*1000]
            {data/hysteresis/joglekar_1.csv};
          \addlegendentry{\SI{110}{\kilo\hertz}}
          \addplot[Maroon, thick, dashed]
            table[col sep=comma, x=V, y expr=\thisrow{I}*1000]
            {data/hysteresis/joglekar_2.csv};
          \addlegendentry{\SI{190}{\kilo\hertz}}
        \end{axis}
      \end{tikzpicture}
    \end{minipage}
  }%
  \subfloat[Biolek]{
    \begin{minipage}[t]{0.32\textwidth}\centering\hspace*{-2mm}
      \begin{tikzpicture}
        \begin{axis}[hyst]
          \addplot[RoyalBlue, thick]
            table[col sep=comma, x=V, y expr=\thisrow{I}*1000]
            {data/hysteresis/biolek_1.csv};
          \addlegendentry{\SI{400}{\kilo\hertz}}
          \addplot[Maroon, thick, dashed]
            table[col sep=comma, x=V, y expr=\thisrow{I}*1000]
            {data/hysteresis/biolek_2.csv};
          \addlegendentry{\SI{900}{\kilo\hertz}}
        \end{axis}
      \end{tikzpicture}
    \end{minipage}
  }

  \subfloat[VTEAM]{
    \begin{minipage}[t]{0.32\textwidth}\centering\hspace*{-2mm}
      \begin{tikzpicture}
        \begin{axis}[hyst]
          \addplot[RoyalBlue, thick]
            table[col sep=comma, x=V, y expr=\thisrow{I}*1000]
            {data/hysteresis/vteam_1.csv};
          \addlegendentry{\SI{10}{\kilo\hertz}}
          \addplot[Maroon, thick, dashed]
            table[col sep=comma, x=V, y expr=\thisrow{I}*1000]
            {data/hysteresis/vteam_2.csv};
          \addlegendentry{\SI{40}{\kilo\hertz}}
        \end{axis}
      \end{tikzpicture}
    \end{minipage}
  }%
  \subfloat[Yakopcic]{
    \begin{minipage}[t]{0.32\textwidth}\centering\hspace*{-2mm}
      \begin{tikzpicture}
        \begin{axis}[hyst]
          \addplot[RoyalBlue, thick]
            table[col sep=comma, x=V, y expr=\thisrow{I}*1000]
            {data/hysteresis/yakopcic_1.csv};
          \addlegendentry{\SI{10}{\hertz}}
          \addplot[Maroon, thick, dashed]
            table[col sep=comma, x=V, y expr=\thisrow{I}*1000]
            {data/hysteresis/yakopcic_2.csv};
          \addlegendentry{\SI{150}{\hertz}}
        \end{axis}
      \end{tikzpicture}
    \end{minipage}
  }%
  \subfloat[MMS]{
    \begin{minipage}[t]{0.32\textwidth}\centering\hspace*{-2mm}
      \begin{tikzpicture}
        \begin{axis}[hyst]
          \addplot[RoyalBlue, thick]
            table[col sep=comma, x=V, y expr=\thisrow{I}*1000]
            {data/hysteresis/mms_1.csv};
          \addlegendentry{\SI{400}{\hertz}}
          \addplot[Maroon, thick, dashed]
            table[col sep=comma, x=V, y expr=\thisrow{I}*1000]
            {data/hysteresis/mms_2.csv};
          \addlegendentry{\SI{1}{\kilo\hertz}}
        \end{axis}
      \end{tikzpicture}
    \end{minipage}
  }%

  \caption{Simulated current-voltage hysteresis loops for the six memristor models at two representative frequencies (solid line = lower frequency; dashed line = higher frequency) under a sinusoidal input voltage with an amplitude of \SI{1}{\volt}.}
  \label{fig:hysteresis-comparison}
\endgroup
\end{figure}

%% file: figures/heatmap-pwm.tex
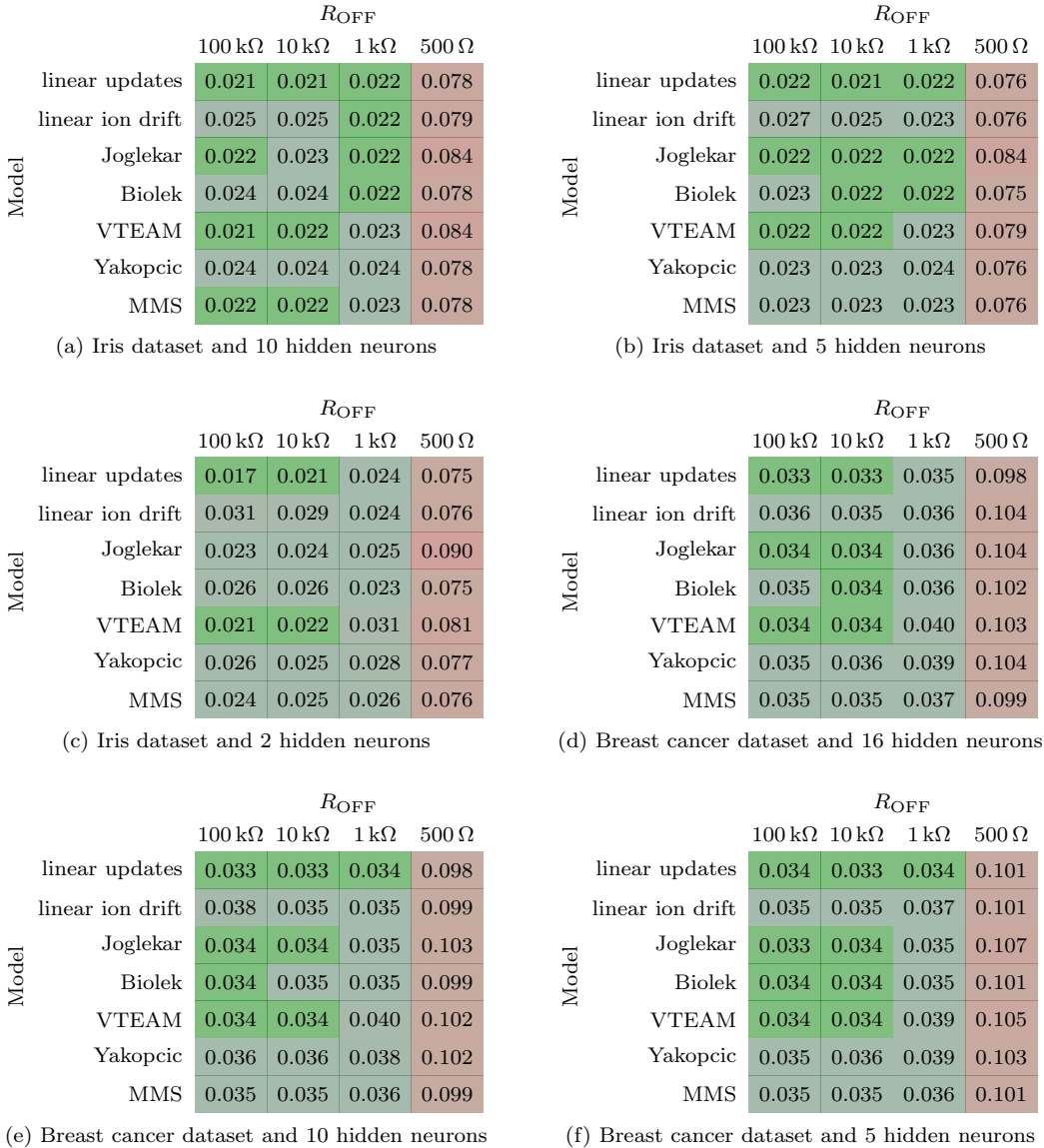
\begin{figure}[!b]
    \centering
    \subfloat[Iris dataset and $10$ hidden neurons\label{fig:pwm-1}]{%
    \begin{minipage}[t]{0.5\textwidth}
        \centering
        \begin{tikzpicture}[font=\scriptsize]
            \node[rotate=90, anchor=south] at (-1.65,-1.9) {Model};
            \node[anchor=south] at (2.5,0.14) {$R_\mathrm{OFF}$};
            \foreach \x [count=\m] in {$\SI{100}{\kilo\ohm}$,$\SI{10}{\kilo\ohm}$,$\SI{1}{\kilo\ohm}$,$\SI{500}{\ohm}$} {
                \node[minimum width=0.95cm, anchor=south] at (\m*0.95, -0.25) {\x};
            }
          
            \foreach \a [count=\i] in {linear updates,linear ion drift,Joglekar,Biolek,VTEAM,Yakopcic,MMS} {
              \node[minimum height=0.495cm, anchor=east] at (0.45,-\i*0.495) {\a};
            }
            
            \foreach \y [count=\n] in {
                {0.021,0.021,0.022,0.078},
                {0.025,0.025,0.022,0.079},
                {0.022,0.023,0.022,0.084},
                {0.024,0.024,0.022,0.078},
                {0.021,0.022,0.023,0.084},
                {0.024,0.024,0.024,0.078},
                {0.022,0.022,0.023,0.078},
              } {      
                \foreach \x [count=\m] in \y {
                  \pgfmathsetmacro{\scaled}{100*((\x < 0.022 ? 0.022 : \x)-0.022)/0.078}
                  \node[fill=Maroon!\scaled!webgreen, fill opacity=0.5, text opacity=1, minimum width=0.95cm, minimum height=0.495cm, inner sep=0pt] at (\m*0.95,-\n*0.495) {$\x$};
                }
              }
          \end{tikzpicture}
    \end{minipage}
    }%
    \subfloat[Iris dataset and $5$ hidden neurons\label{fig:pwm-2}]{%
    \begin{minipage}[t]{0.5\textwidth}
        \centering
        \begin{tikzpicture}[font=\scriptsize]
            \node[rotate=90, anchor=south] at (-1.65,-1.9) {Model};
            \node[anchor=south] at (2.5,0.14) {$R_\mathrm{OFF}$};
            \foreach \x [count=\m] in {$\SI{100}{\kilo\ohm}$,$\SI{10}{\kilo\ohm}$,$\SI{1}{\kilo\ohm}$,$\SI{500}{\ohm}$} {
              \node[minimum width=0.95cm, anchor=south] at (\m*0.95, -0.25) {\x};
            }
          
            \foreach \a [count=\i] in {linear updates,linear ion drift,Joglekar,Biolek,VTEAM,Yakopcic,MMS} {
                \node[minimum height=0.495cm, anchor=east] at (0.45,-\i*0.495) {\a};
            }
            
            \foreach \y [count=\n] in {
                {0.022,0.021,0.022,0.076},
                {0.027,0.025,0.023,0.076},
                {0.022,0.022,0.022,0.084},
                {0.023,0.022,0.022,0.075},
                {0.022,0.022,0.023,0.079},
                {0.023,0.023,0.024,0.076},
                {0.023,0.023,0.023,0.076},
              } {      
                \foreach \x [count=\m] in \y {
                  \pgfmathsetmacro{\scaled}{100*((\x < 0.022 ? 0.022 : \x)-0.022)/0.078}
                  \node[fill=Maroon!\scaled!webgreen, fill opacity=0.5, text opacity=1, minimum width=0.95cm, minimum height=0.495cm, inner sep=0pt] at (\m*0.95,-\n*0.495) {$\x$};
                }
              }
          \end{tikzpicture}
    \end{minipage}
    }%
    
    \subfloat[Iris dataset and $2$ hidden neurons\label{fig:pwm-3}]{%
    \begin{minipage}[t]{0.5\textwidth}
        \centering
        \begin{tikzpicture}[font=\scriptsize]
            \node[rotate=90, anchor=south] at (-1.65,-1.9) {Model};
            \node[anchor=south] at (2.5,0.14) {$R_\mathrm{OFF}$};
            \foreach \x [count=\m] in {$\SI{100}{\kilo\ohm}$,$\SI{10}{\kilo\ohm}$,$\SI{1}{\kilo\ohm}$,$\SI{500}{\ohm}$} {
                \node[minimum width=0.95cm, anchor=south] at (\m*0.95, -0.25) {\x};
            }
          
            \foreach \a [count=\i] in {linear updates,linear ion drift,Joglekar,Biolek,VTEAM,Yakopcic,MMS} {
              \node[minimum height=0.495cm, anchor=east] at (0.45,-\i*0.495) {\a};
            }
            
            \foreach \y [count=\n] in {
                {0.017,0.021,0.024,0.075},
                {0.031,0.029,0.024,0.076},
                {0.023,0.024,0.025,0.090},
                {0.026,0.026,0.023,0.075},
                {0.021,0.022,0.031,0.081},
                {0.026,0.025,0.028,0.077},
                {0.024,0.025,0.026,0.076},
              } {      
                \foreach \x [count=\m] in \y {
                  \pgfmathsetmacro{\scaled}{100*((\x < 0.022 ? 0.022 : \x)-0.022)/0.078}
                  \node[fill=Maroon!\scaled!webgreen, fill opacity=0.5, text opacity=1, minimum width=0.95cm, minimum height=0.495cm, inner sep=0pt] at (\m*0.95,-\n*0.495) {$\x$};
                }
              }
          \end{tikzpicture}
    \end{minipage}
    }%
    \subfloat[Breast cancer dataset and $16$ hidden neurons\label{fig:pwm-4}]{%
    \begin{minipage}[t]{0.5\textwidth}
        \centering
        \begin{tikzpicture}[font=\scriptsize]
            \node[rotate=90, anchor=south] at (-1.65,-1.9) {Model};
            \node[anchor=south] at (2.5,0.14) {$R_\mathrm{OFF}$};
            \foreach \x [count=\m] in {$\SI{100}{\kilo\ohm}$,$\SI{10}{\kilo\ohm}$,$\SI{1}{\kilo\ohm}$,$\SI{500}{\ohm}$} {
                \node[minimum width=0.95cm, anchor=south] at (\m*0.95, -0.25) {\x};
            }
          
            \foreach \a [count=\i] in {linear updates,linear ion drift,Joglekar,Biolek,VTEAM,Yakopcic,MMS} {
              \node[minimum height=0.495cm, anchor=east] at (0.45,-\i*0.495) {\a};
            }
            
            \foreach \y [count=\n] in {
                {0.033,0.033,0.035,0.098},
                {0.036,0.035,0.036,0.104},
                {0.034,0.034,0.036,0.104},
                {0.035,0.034,0.036,0.102},
                {0.034,0.034,0.040,0.103},
                {0.035,0.036,0.039,0.104},
                {0.035,0.035,0.037,0.099},
              } {      
                \foreach \x [count=\m] in \y {
                  \pgfmathsetmacro{\scaled}{100*((\x < 0.034 ? 0.034 : \x)-0.034)/0.100}
                  \node[fill=Maroon!\scaled!webgreen, fill opacity=0.5, text opacity=1, minimum width=0.95cm, minimum height=0.495cm, inner sep=0pt] at (\m*0.95,-\n*0.495) {$\x$};
                }
              }
          \end{tikzpicture}
    \end{minipage}
    }%

    \subfloat[Breast cancer dataset and $10$ hidden neurons\label{fig:pwm-5}]{%
    \begin{minipage}[t]{0.5\textwidth}
        \centering
        \begin{tikzpicture}[font=\scriptsize]
            \node[rotate=90, anchor=south] at (-1.65,-1.9) {Model};
            \node[anchor=south] at (2.5,0.14) {$R_\mathrm{OFF}$};
            \foreach \x [count=\m] in {$\SI{100}{\kilo\ohm}$,$\SI{10}{\kilo\ohm}$,$\SI{1}{\kilo\ohm}$,$\SI{500}{\ohm}$} {
                \node[minimum width=0.95cm, anchor=south] at (\m*0.95, -0.25) {\x};
            }
          
            \foreach \a [count=\i] in {linear updates,linear ion drift,Joglekar,Biolek,VTEAM,Yakopcic,MMS} {
              \node[minimum height=0.495cm, anchor=east] at (0.45,-\i*0.495) {\a};
            }
            
            \foreach \y [count=\n] in {
                {0.033,0.033,0.034,0.098},
                {0.038,0.035,0.035,0.099},
                {0.034,0.034,0.035,0.103},
                {0.034,0.035,0.035,0.099},
                {0.034,0.034,0.040,0.102},
                {0.036,0.036,0.038,0.102},
                {0.035,0.035,0.036,0.099},
              } {      
                \foreach \x [count=\m] in \y {
                  \pgfmathsetmacro{\scaled}{100*((\x < 0.034 ? 0.034 : \x)-0.034)/0.100}
                  \node[fill=Maroon!\scaled!webgreen, fill opacity=0.5, text opacity=1, minimum width=0.95cm, minimum height=0.495cm, inner sep=0pt] at (\m*0.95,-\n*0.495) {$\x$};
                }
              }
          \end{tikzpicture}
    \end{minipage}
    }%
    \subfloat[Breast cancer dataset and $5$ hidden neurons\label{fig:pwm-6}]{%
    \begin{minipage}[t]{0.5\textwidth}
        \centering
        \begin{tikzpicture}[font=\scriptsize]
            \node[rotate=90, anchor=south] at (-1.65,-1.9) {Model};
            \node[anchor=south] at (2.5,0.14) {$R_\mathrm{OFF}$};
            \foreach \x [count=\m] in {$\SI{100}{\kilo\ohm}$,$\SI{10}{\kilo\ohm}$,$\SI{1}{\kilo\ohm}$,$\SI{500}{\ohm}$} {
                \node[minimum width=0.95cm, anchor=south] at (\m*0.95, -0.25) {\x};
            }
          
            \foreach \a [count=\i] in {linear updates,linear ion drift,Joglekar,Biolek,VTEAM,Yakopcic,MMS} {
              \node[minimum height=0.495cm, anchor=east] at (0.45,-\i*0.495) {\a};
            }
            
            \foreach \y [count=\n] in {
                {0.034,0.033,0.034,0.101},
                {0.035,0.035,0.037,0.101},
                {0.033,0.034,0.035,0.107},
                {0.034,0.034,0.035,0.101},
                {0.034,0.034,0.039,0.105},
                {0.035,0.036,0.039,0.103},
                {0.035,0.035,0.036,0.101},
              } {      
                \foreach \x [count=\m] in \y {
                  \pgfmathsetmacro{\scaled}{100*((\x < 0.034 ? 0.034 : \x)-0.034)/0.100}
                  \node[fill=Maroon!\scaled!webgreen, fill opacity=0.5, text opacity=1, minimum width=0.95cm, minimum height=0.495cm, inner sep=0pt] at (\m*0.95,-\n*0.495) {$\x$};
                }
              }
          \end{tikzpicture}
    \end{minipage}
    }%
    \caption{Heatmaps of minimum losses of EqProp simulations with PWM.}
    \label{fig:heatmap-pwm}
\end{figure}

%% file: figures/heatmap-pam.tex
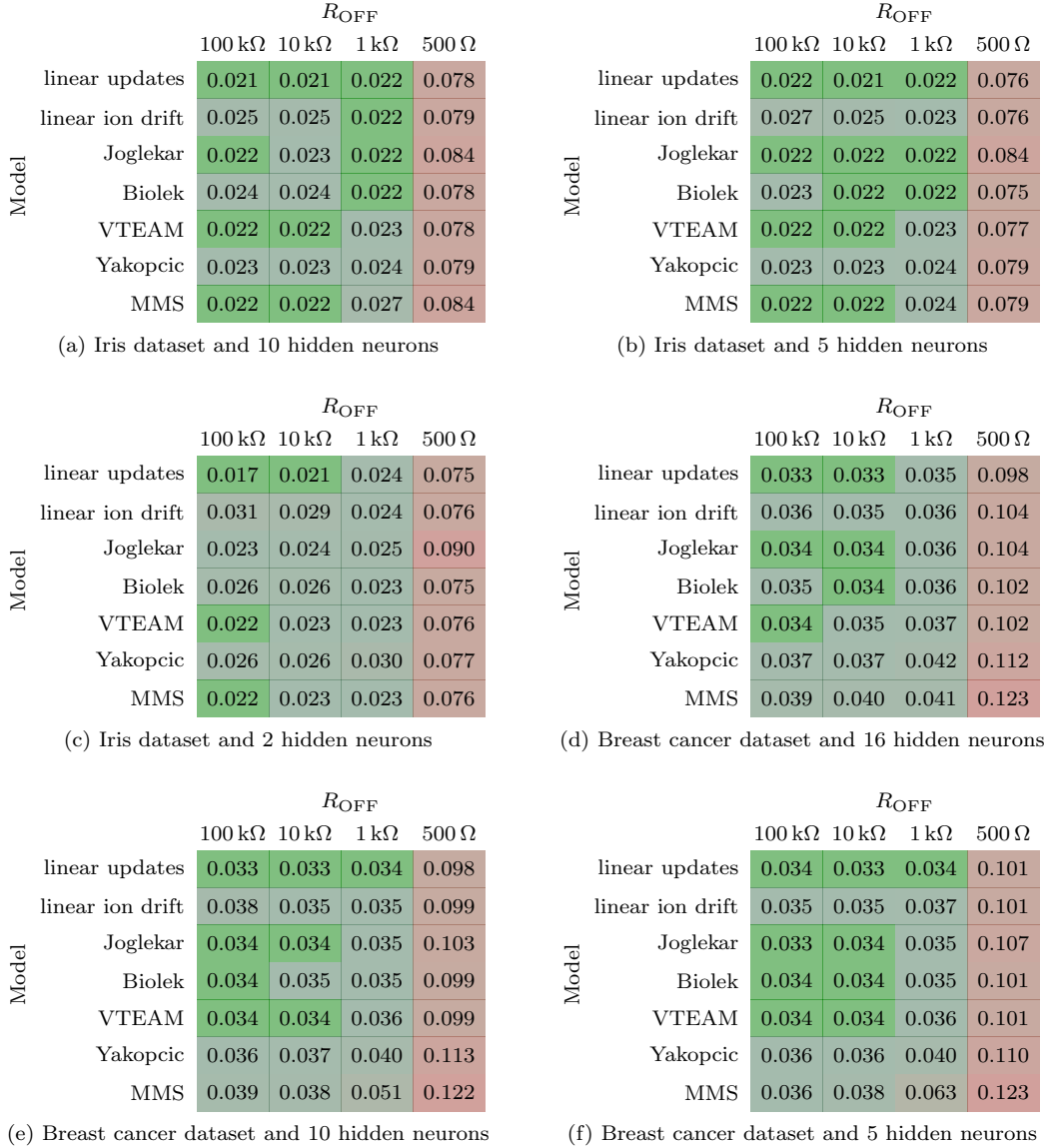
\begin{figure}[!t]
    \centering
    \subfloat[Iris dataset and $10$ hidden neurons\label{fig:pam-1}]{%
    \begin{minipage}[t]{0.5\textwidth}
        \centering
        \begin{tikzpicture}[font=\scriptsize]
            \node[rotate=90, anchor=south] at (-1.65,-1.9) {Model};
            \node[anchor=south] at (2.5,0.14) {$R_\mathrm{{OFF}}$};
            \foreach \x [count=\m] in {$\SI{100}{\kilo\ohm}$,$\SI{10}{\kilo\ohm}$,$\SI{1}{\kilo\ohm}$,$\SI{500}{\ohm}$} {
                \node[minimum width=0.95cm, anchor=south] at (\m*0.95, -0.25) {\x};
            }
          
            \foreach \a [count=\i] in {linear updates,linear ion drift,Joglekar,Biolek,VTEAM,Yakopcic,MMS} {
              \node[minimum height=0.495cm, anchor=east] at (0.45,-\i*0.495) {\a};
            }
            
            \foreach \y [count=\n] in {
                {0.021,0.021,0.022,0.078},
                {0.025,0.025,0.022,0.079},
                {0.022,0.023,0.022,0.084},
                {0.024,0.024,0.022,0.078},
                {0.022,0.022,0.023,0.078},
                {0.023,0.023,0.024,0.079},
                {0.022,0.022,0.027,0.084},
              } {      
                \foreach \x [count=\m] in \y {
                  \pgfmathsetmacro{\scaled}{100*((\x < 0.022 ? 0.022 : \x)-0.022)/0.078}
                  \node[fill=Maroon!\scaled!webgreen, fill opacity=0.5, text opacity=1, minimum width=0.95cm, minimum height=0.495cm, inner sep=0pt] at (\m*0.95,-\n*0.495) {$\x$};
                }
              }
          \end{tikzpicture}
    \end{minipage}
    }%
    \subfloat[Iris dataset and $5$ hidden neurons\label{fig:pam-2}]{%
    \begin{minipage}[t]{0.5\textwidth}
        \centering
        \begin{tikzpicture}[font=\scriptsize]
            \node[rotate=90, anchor=south] at (-1.65,-1.9) {Model};
            \node[anchor=south] at (2.5,0.14) {$R_\mathrm{{OFF}}$};
            \foreach \x [count=\m] in {$\SI{100}{\kilo\ohm}$,$\SI{10}{\kilo\ohm}$,$\SI{1}{\kilo\ohm}$,$\SI{500}{\ohm}$} {
              \node[minimum width=0.95cm, anchor=south] at (\m*0.95, -0.25) {\x};
            }
          
            \foreach \a [count=\i] in {linear updates,linear ion drift,Joglekar,Biolek,VTEAM,Yakopcic,MMS} {
                \node[minimum height=0.495cm, anchor=east] at (0.45,-\i*0.495) {\a};
            }
            
            \foreach \y [count=\n] in {
                {0.022,0.021,0.022,0.076},
                {0.027,0.025,0.023,0.076},
                {0.022,0.022,0.022,0.084},
                {0.023,0.022,0.022,0.075},
                {0.022,0.022,0.023,0.077},
                {0.023,0.023,0.024,0.079},
                {0.022,0.022,0.024,0.079},
              } {      
                \foreach \x [count=\m] in \y {
                  \pgfmathsetmacro{\scaled}{100*((\x < 0.022 ? 0.022 : \x)-0.022)/0.078}
                  \node[fill=Maroon!\scaled!webgreen, fill opacity=0.5, text opacity=1, minimum width=0.95cm, minimum height=0.495cm, inner sep=0pt] at (\m*0.95,-\n*0.495) {$\x$};
                }
              }
          \end{tikzpicture}
    \end{minipage}
    }%

    \subfloat[Iris dataset and $2$ hidden neurons\label{fig:pam-3}]{%
    \begin{minipage}[t]{0.5\textwidth}
        \centering
        \begin{tikzpicture}[font=\scriptsize]
            \node[rotate=90, anchor=south] at (-1.65,-1.9) {Model};
            \node[anchor=south] at (2.5,0.14) {$R_\mathrm{{OFF}}$};
            \foreach \x [count=\m] in {$\SI{100}{\kilo\ohm}$,$\SI{10}{\kilo\ohm}$,$\SI{1}{\kilo\ohm}$,$\SI{500}{\ohm}$} {
                \node[minimum width=0.95cm, anchor=south] at (\m*0.95, -0.25) {\x};
            }
          
            \foreach \a [count=\i] in {linear updates,linear ion drift,Joglekar,Biolek,VTEAM,Yakopcic,MMS} {
              \node[minimum height=0.495cm, anchor=east] at (0.45,-\i*0.495) {\a};
            }
            
            \foreach \y [count=\n] in {
                {0.017,0.021,0.024,0.075},
                {0.031,0.029,0.024,0.076},
                {0.023,0.024,0.025,0.090},
                {0.026,0.026,0.023,0.075},
                {0.022,0.023,0.023,0.076},
                {0.026,0.026,0.030,0.077},
                {0.022,0.023,0.023,0.076},
              } {      
                \foreach \x [count=\m] in \y {
                  \pgfmathsetmacro{\scaled}{100*((\x < 0.022 ? 0.022 : \x)-0.022)/0.078}
                  \node[fill=Maroon!\scaled!webgreen, fill opacity=0.5, text opacity=1, minimum width=0.95cm, minimum height=0.495cm, inner sep=0pt] at (\m*0.95,-\n*0.495) {$\x$};
                }
              }
          \end{tikzpicture}
    \end{minipage}
    }%
    \subfloat[Breast cancer dataset and $16$ hidden neurons\label{fig:pam-4}]{%
    \begin{minipage}[t]{0.5\textwidth}
        \centering
        \begin{tikzpicture}[font=\scriptsize]
            \node[rotate=90, anchor=south] at (-1.65,-1.9) {Model};
            \node[anchor=south] at (2.5,0.14) {$R_\mathrm{{OFF}}$};
            \foreach \x [count=\m] in {$\SI{100}{\kilo\ohm}$,$\SI{10}{\kilo\ohm}$,$\SI{1}{\kilo\ohm}$,$\SI{500}{\ohm}$} {
                \node[minimum width=0.95cm, anchor=south] at (\m*0.95, -0.25) {\x};
            }
          
            \foreach \a [count=\i] in {linear updates,linear ion drift,Joglekar,Biolek,VTEAM,Yakopcic,MMS} {
              \node[minimum height=0.495cm, anchor=east] at (0.45,-\i*0.495) {\a};
            }
            
            \foreach \y [count=\n] in {
                {0.033,0.033,0.035,0.098},
                {0.036,0.035,0.036,0.104},
                {0.034,0.034,0.036,0.104},
                {0.035,0.034,0.036,0.102},
                {0.034,0.035,0.037,0.102},
                {0.037,0.037,0.042,0.112},
                {0.039,0.040,0.041,0.123},
              } {      
                \foreach \x [count=\m] in \y {
                  \pgfmathsetmacro{\scaled}{100*((\x < 0.034 ? 0.034 : \x)-0.034)/0.100}
                  \node[fill=Maroon!\scaled!webgreen, fill opacity=0.5, text opacity=1, minimum width=0.95cm, minimum height=0.495cm, inner sep=0pt] at (\m*0.95,-\n*0.495) {$\x$};
                }
              }
          \end{tikzpicture}
    \end{minipage}
    }%
    
    \subfloat[Breast cancer dataset and $10$ hidden neurons\label{fig:pam-5}]{%
    \begin{minipage}[t]{0.5\textwidth}
        \centering
        \begin{tikzpicture}[font=\scriptsize]
            \node[rotate=90, anchor=south] at (-1.65,-1.9) {Model};
            \node[anchor=south] at (2.5,0.14) {$R_\mathrm{{OFF}}$};
            \foreach \x [count=\m] in {$\SI{100}{\kilo\ohm}$,$\SI{10}{\kilo\ohm}$,$\SI{1}{\kilo\ohm}$,$\SI{500}{\ohm}$} {
                \node[minimum width=0.95cm, anchor=south] at (\m*0.95, -0.25) {\x};
            }
          
            \foreach \a [count=\i] in {linear updates,linear ion drift,Joglekar,Biolek,VTEAM,Yakopcic,MMS} {
              \node[minimum height=0.495cm, anchor=east] at (0.45,-\i*0.495) {\a};
            }
            
            \foreach \y [count=\n] in {
                {0.033,0.033,0.034,0.098},
                {0.038,0.035,0.035,0.099},
                {0.034,0.034,0.035,0.103},
                {0.034,0.035,0.035,0.099},
                {0.034,0.034,0.036,0.099},
                {0.036,0.037,0.040,0.113},
                {0.039,0.038,0.051,0.122},
              } {      
                \foreach \x [count=\m] in \y {
                  \pgfmathsetmacro{\scaled}{100*((\x < 0.034 ? 0.034 : \x)-0.034)/0.100}
                  \node[fill=Maroon!\scaled!webgreen, fill opacity=0.5, text opacity=1, minimum width=0.95cm, minimum height=0.495cm, inner sep=0pt] at (\m*0.95,-\n*0.495) {$\x$};
                }
              }
          \end{tikzpicture}
    \end{minipage}
    }%
    \subfloat[Breast cancer dataset and $5$ hidden neurons\label{fig:pam-6}]{%
    \begin{minipage}[t]{0.5\textwidth}
        \centering
        \begin{tikzpicture}[font=\scriptsize]
            \node[rotate=90, anchor=south] at (-1.65,-1.9) {Model};
            \node[anchor=south] at (2.5,0.14) {$R_\mathrm{{OFF}}$};
            \foreach \x [count=\m] in {$\SI{100}{\kilo\ohm}$,$\SI{10}{\kilo\ohm}$,$\SI{1}{\kilo\ohm}$,$\SI{500}{\ohm}$} {
                \node[minimum width=0.95cm, anchor=south] at (\m*0.95, -0.25) {\x};
            }
          
            \foreach \a [count=\i] in {linear updates,linear ion drift,Joglekar,Biolek,VTEAM,Yakopcic,MMS} {
              \node[minimum height=0.495cm, anchor=east] at (0.45,-\i*0.495) {\a};
            }
            
            \foreach \y [count=\n] in {
                {0.034,0.033,0.034,0.101},
                {0.035,0.035,0.037,0.101},
                {0.033,0.034,0.035,0.107},
                {0.034,0.034,0.035,0.101},
                {0.034,0.034,0.036,0.101},
                {0.036,0.036,0.040,0.110},
                {0.036,0.038,0.063,0.123},
              } {      
                \foreach \x [count=\m] in \y {
                  \pgfmathsetmacro{\scaled}{100*((\x < 0.034 ? 0.034 : \x)-0.034)/0.100}
                  \node[fill=Maroon!\scaled!webgreen, fill opacity=0.5, text opacity=1, minimum width=0.95cm, minimum height=0.495cm, inner sep=0pt] at (\m*0.95,-\n*0.495) {$\x$};
                }
              }
          \end{tikzpicture}
    \end{minipage}
    }%
    \caption{Heatmaps of minimum losses of EqProp simulations with PAM.}
    \label{fig:heatmap-pam}
\end{figure}

%% file: figures/sneak-path.tex
\begin{figure}[!h]
    \centering
    \begin{circuitikz}[scale=0.83, transform shape, font=\large]
        \def\cols{2}
        \def\rows{2}
        \def\colsep{2}
        \def\rowsep{1.8}
        \def\memleft{1.25}
        \def\memdown{1.25}
        
        \foreach \row in {1,...,\rows} {
            \foreach \col in {1,...,\cols} {
                \pgfmathsetmacro{\xstart}{(\col-\memleft)*\colsep}
                \pgfmathsetmacro{\ystart}{(\rows*\rowsep)-((\row-1)*\rowsep)-1}
                \pgfmathsetmacro{\xend}{\xstart+\memleft}
                \pgfmathsetmacro{\yend}{\ystart-\memdown}
                
                \ifnum \col=\cols
                    \ifnum \row=1
                        \draw (\xstart,\ystart) to[memristor, -, l_=$G_{\row,\col}$] (\xend,\yend);
                        \draw[Maroon, thick, dashed] (\xstart-0.05,\ystart-0.1) -- (\xend-0.1,\yend-0.05);
                    \else
                        \draw (\xstart,\ystart) to[memristor, -*, l_=$G_{\row,\col}$] (\xend,\yend);
                        \draw[Maroon, thick, dashed] (\xstart+0.05,\ystart+0.1) -- (\xend-0.1,\yend+0.25);
                    \fi
                \else 
                    \ifnum \row=1
                        \draw (\xstart,\ystart) to[memristor, *-, l_=$G_{\row,\col}$] (\xend,\yend);
                        \draw[RoyalBlue, thick] (\xstart+0.05,\ystart+0.1) -- (\xend+0.1,\yend+0.055);
                    \else
                        \draw (\xstart,\ystart) to[memristor, *-*, l_=$G_{\row,\col}$] (\xend,\yend);
                        \draw[Maroon, thick, dashed] (\xstart-0.2,\ystart+0.05) -- (\xend-0.1,\yend-0.05);
                    \fi
                \fi
            }
        }

        \foreach \row in {1,...,\rows} {
            \pgfmathsetmacro{\ystart}{(\rows*\rowsep)-((\row-1)*\rowsep)-1}
            \draw ({-(\colsep/2)-0.2},\ystart) -- ({(1-\memleft)*\colsep},\ystart);
            \ifnum \row=1
                \draw[RoyalBlue, thick] ({-(\colsep/2)-0.2},\ystart+0.1) -- ({(1-\memleft)*\colsep+0.05},\ystart+0.1);
                \draw[Maroon, thick, dashed] ({-(\colsep/2)-0.2},\ystart-0.1) -- ({(1-\memleft)*\colsep+\colsep-0.05},\ystart-0.1);
            \fi
        }
        
        \foreach \col in {1,...,\cols} {
            \pgfmathsetmacro{\x}{(\col-\memleft)*\colsep+\memleft}
            \pgfmathsetmacro{\y}{\rowsep-\memdown-1}
            \draw (\x,{(\rows)*\rowsep-\memdown-1}) -- (\x,\y-\memdown*0.5);
            \ifnum \col=1
                \draw[RoyalBlue, thick, -{Stealth[scale=0.94]}] (\x+0.1,{(\rows)*\rowsep-\memdown-0.95}) -- (\x+0.1,\y-\memdown*0.6-0.05);
                \draw[RoyalBlue, thick, -{Stealth[scale=0.94]}] (\x+0.1,{(\rows)*\rowsep-\memdown-0.95}) -- (\x+0.1,\y-\memdown*0.6-0.05);
                \draw[Maroon, thick, dashed, -{Stealth[scale=0.94]}] (\x-0.1,{(\rows-1)*\rowsep-\memdown-1.05}) -- (\x-0.1,\y-\memdown*0.6-0.05);
            \fi
            \ifnum \col=2
                \draw[Maroon, thick, dashed] (\x-0.1,{(\rows)*\rowsep-\memdown-1.05}) -- (\x-0.1,\y+0.25);
            \fi
        }

        \draw ({(1-\memleft)*\colsep},{\rows*\rowsep-1}) -- ({(\cols-\memleft)*\colsep},{\rows*\rowsep-1});
        
        \pgfmathsetmacro{\ymax}{(\rows-1)*\rowsep}
        \foreach \row in {2,...,\rows} {
            \foreach \col in {2,...,\cols} {
                \pgfmathsetmacro{\x}{(\col-1-\memleft)*\colsep}
                \pgfmathsetmacro{\y}{(\row-1)*\rowsep-1}
                \draw (\x,\y) -- ({\x+\memleft-0.3},\y) to[crossing, bipoles/crossing/size=0.4] ({\x+\memleft+0.3},\y) -- ({\x+\colsep},\y);
                \draw[Maroon, thick, dashed] (\x-0.15,\y+0.1) -- ({\x+\colsep+0.05},\y+0.1);
            }
        }
        
        \begin{scope}[shift={(3.5,2)}]
            \draw[RoyalBlue, thick, -{Stealth[scale=0.94]}] (0,-0.5) -- (0.5,-0.5) node[right, text=black]{\small desired path};
            \draw[Maroon, thick, dashed, -{Stealth[scale=0.94]}] (0,-1) -- (0.5,-1) node[right, text=black]{\small sneak path};
        \end{scope}
    \end{circuitikz}
    \caption{Sneak-path current in a $2 \times 2$ crossbar array~\cite{humood_high-density_2019}.}
    \label{fig:sneak-path}
\end{figure}
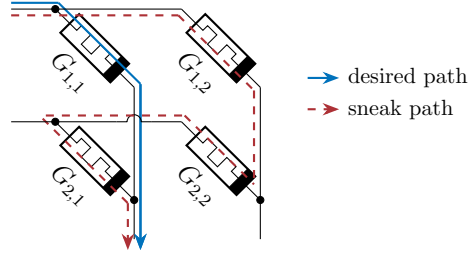

%% file: figures/crossbar-line-resistances.tex
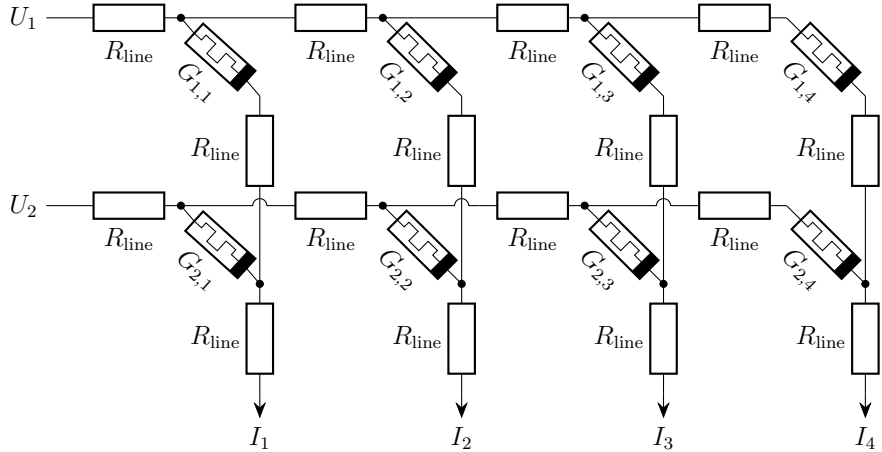
\begin{figure}[!t]
    \centering
    \begin{circuitikz}[scale=0.83, transform shape, font=\large]
        \tikzset{small resistor/.style={resistor, transform shape, scale=0.5, european}}
        \def\cols{4}
        \def\rows{2}
        \def\colsep{3.2}
        \def\rowsep{3}
        \def\memleft{1.25}
        \def\memdown{1.25}
        
        \foreach \row in {1,...,\rows} {
            \foreach \col in {1,...,\cols} {
                \pgfmathsetmacro{\xstart}{(\col-\memleft)*\colsep}
                \pgfmathsetmacro{\ystart}{(\rows*\rowsep)-((\row-1)*\rowsep)-1}
                \pgfmathsetmacro{\xend}{\xstart+\memleft}
                \pgfmathsetmacro{\yend}{\ystart-\memdown}
                
                \ifnum \col=\cols
                    \ifnum \row=1
                        \draw (\xstart,\ystart) to[memristor, -, l_=$G_{\row,\col}$] (\xend,\yend);
                    \else
                        \draw (\xstart,\ystart) to[memristor, -*, l_=$G_{\row,\col}$] (\xend,\yend);
                    \fi
                \else 
                    \ifnum \row=1
                        \draw (\xstart,\ystart) to[memristor, *-, l_=$G_{\row,\col}$] (\xend,\yend);
                    \else
                        \draw (\xstart,\ystart) to[memristor, *-*, l_=$G_{\row,\col}$] (\xend,\yend);
                    \fi
                \fi
            }
        }
        
        \foreach \col in {1,...,\cols} {
            \pgfmathsetmacro{\x}{(\col-\memleft)*\colsep+\memleft}
            \foreach \row in {1,...,\rows} {
                \pgfmathsetmacro{\y}{(\row)*\rowsep-1-\memdown}
                \ifnum \row=1
                    \draw (\x,\y) to[resistor, european, l_=$R_\mathrm{line}$] (\x,\y-\rowsep+\memdown);
                    \draw[-{Stealth[scale=1.3]}] (\x,\y-\rowsep+\memdown) -- (\x,\y-\rowsep+\memdown-0.4);
                    \node[below] at (\x,\y-\rowsep+\memdown-0.4) {$I_{\col}$};
                \else
                    \draw (\x,\y) to[resistor, european, l_=$R_\mathrm{line}$] (\x,\y-\rowsep+\memdown) -- (\x,\y-\rowsep);
                \fi
            }
        }
        
        \pgfmathsetmacro{\ymax}{(\rows-1)*\rowsep}
        \foreach \row in {1,...,\rows} {
            \foreach \col in {1,...,\cols} {
                \pgfmathsetmacro{\x}{(\col-1-\memleft)*\colsep}
                \pgfmathsetmacro{\y}{(\row)*\rowsep-1}
                \ifnum \col=1
                    \pgfmathtruncatemacro{\rownr}{\rows-\row+1}
                    \node[left] at (\x+\colsep/3,\y) {$U_{\rownr}$};
                    \draw (\x+\colsep/3,\y) -- ({\x+\memleft+0.3},\y) to[resistor, european, l_=$R_\mathrm{line}$] ({\x+\colsep},\y);
                \else
                    \ifnum \row=\rows
                        \draw (\x,\y) -- ({\x+\memleft+0.3},\y) to[resistor, european, l_=$R_\mathrm{line}$] ({\x+\colsep},\y);
                    \else
                        \draw (\x,\y) -- ({\x+\memleft-0.3},\y) to[crossing, bipoles/crossing/size=0.4] ({\x+\memleft+0.3},\y) to[resistor, european, l_=$R_\mathrm{line}$] ({\x+\colsep},\y);
                    \fi
                \fi
            }
        }
    \end{circuitikz}
    \caption{Memristor crossbar array with parasitic line resistances~\cite{jeong_parasitic_2018, nguyen_quantization_2022}.}
    \label{fig:crossbar-line-resistances}
\end{figure}